\documentclass{article}

    \PassOptionsToPackage{numbers}{natbib}
\newcommand{\citew}[1]{[\citenum{#1}]}
\newcommand{\citeay}[1]{\citeauthor{#1}, \citeyear{#1}}

\usepackage{etoolbox}
\newtoggle{arxiv}
\toggletrue{arxiv}   

\iftoggle{arxiv}{
  \usepackage[preprint]{neurips_2025}
}{
  \usepackage[final]{neurips_2025}
}

\usepackage[utf8]{inputenc} 
\usepackage[T1]{fontenc}    
\usepackage{hyperref}       
\usepackage{url}            
\usepackage{booktabs}       
\usepackage{amsfonts}       
\usepackage{nicefrac}       
\usepackage{microtype}      
\usepackage{xcolor}         
\usepackage{amssymb}
\usepackage{amsmath}
\usepackage{algorithm}
\usepackage{algpseudocode}
\usepackage{cleveref}
\usepackage{amsfonts}
\usepackage{wrapfig}
\usepackage{booktabs}  
\usepackage{graphicx}
\graphicspath{{figs/}}
\title{Self-diffusion for Solving Inverse Problems}
\hypersetup{
    colorlinks=false,
    pdfborder={0 0 0}
}
\usepackage{enumitem}

\author{%
  Guanxiong Luo \\
   \And
   Shoujin Huang \\
  \iftoggle{arxiv}{
  \And Yanlong Yang\thanks{Yang contributed to the development of the radar angle estimation recovery. However, in accordance with NeurIPS 2025 authorship policies, which do not allow adding authors after abstract submission, his name could not be included in the official conference version of this paper. We sincerely apologize for this oversight and gratefully acknowledge his contribution.}\\
  }{}
}

\begin{document}

\maketitle

\begin{abstract}
    We propose \emph{self-diffusion}, a novel framework for solving inverse problems without relying on pretrained generative models. Traditional diffusion-based approaches require training a model on a clean dataset to learn to reverse the forward noising process. This model is then used to sample clean solutions---corresponding to posterior sampling from a Bayesian perspective---that are consistent with the observed data under a specific task. In contrast, self-diffusion introduces a self-contained iterative process that alternates between noising and denoising steps to progressively refine its estimate of the solution. At each step of self-diffusion, noise is added to the current estimate, and a self-denoiser, which is a single untrained convolutional network randomly initialized from scratch, is continuously trained for certain iterations via a data fidelity loss to predict the solution from the noisy estimate. Essentially, self-diffusion exploits the spectral bias of neural networks and modulates it through a scheduled noise process. Without relying on pretrained score functions or external denoisers, this approach still remains adaptive to arbitrary forward operators and noisy observations, making it highly flexible and broadly applicable. We demonstrate the effectiveness of our approach on a variety of linear inverse problems, showing that self-diffusion achieves competitive or superior performance compared to other methods.
\end{abstract}

\section{Introduction}
\label{sec:contributions}
Inverse problems involve reconstructing a hidden cause---often called the \emph{source} e.g., an image or signal---from observed data. The observed data is often incomplete and noisy, which leads to ill-posed problems. These problems occur in a wide range of domains and are common in fields like signal and image processing, including tasks like medical imaging reconstruction, image denoising, and low-level vision.
Prior knowledge about the source is particularly valuable when addressing ill-posed problems, where observed data alone may not be sufficient for an accurate and faithful reconstruction. A Bayesian framework offers an approach in such scenarios by combining observed data with prior information. In this view, the goal is to infer the underlying source given the observations by estimating its posterior distribution. This posterior reflects a balance between the likelihood, which is determined by the forward model and noise, and the prior, which encodes assumptions or knowledge about the source. Because the likelihood is often fixed by the physical model, the choice of prior plays a critical role in improving the quality and stability of the reconstruction.

The conventional approach often relies on handcrafted priors to regularize the solution space and steer optimization toward more plausible outcomes. Classic techniques such as Tikhonov regularization \citew{tikhonov1977solutions}, total variation \citew{Rudin1992}, and sparsity-promoting methods \citew{Donoho2006, Candes2006} have been widely used. While these methods can be effective in certain settings, they often struggle to faithfully recover complex signals when the observed data is insufficient.
Thereafter, supervised learning approaches, since deep neural networks have gained popularity, are used to learn mappings from large datasets of input-output pairs \citew{Jin2017, Adler2017}. These learned models have demonstrated strong performance in specific tasks. However, their effectiveness typically depends on the availability of large amounts of labeled training data, which may not always be accessible. Furthermore, these models often lack robustness to distribution shifts, such as varying noise levels or different acquisition settings, limiting their practical utility in some situations.
\begin{figure}
    \centering
    \includegraphics[width=\textwidth]{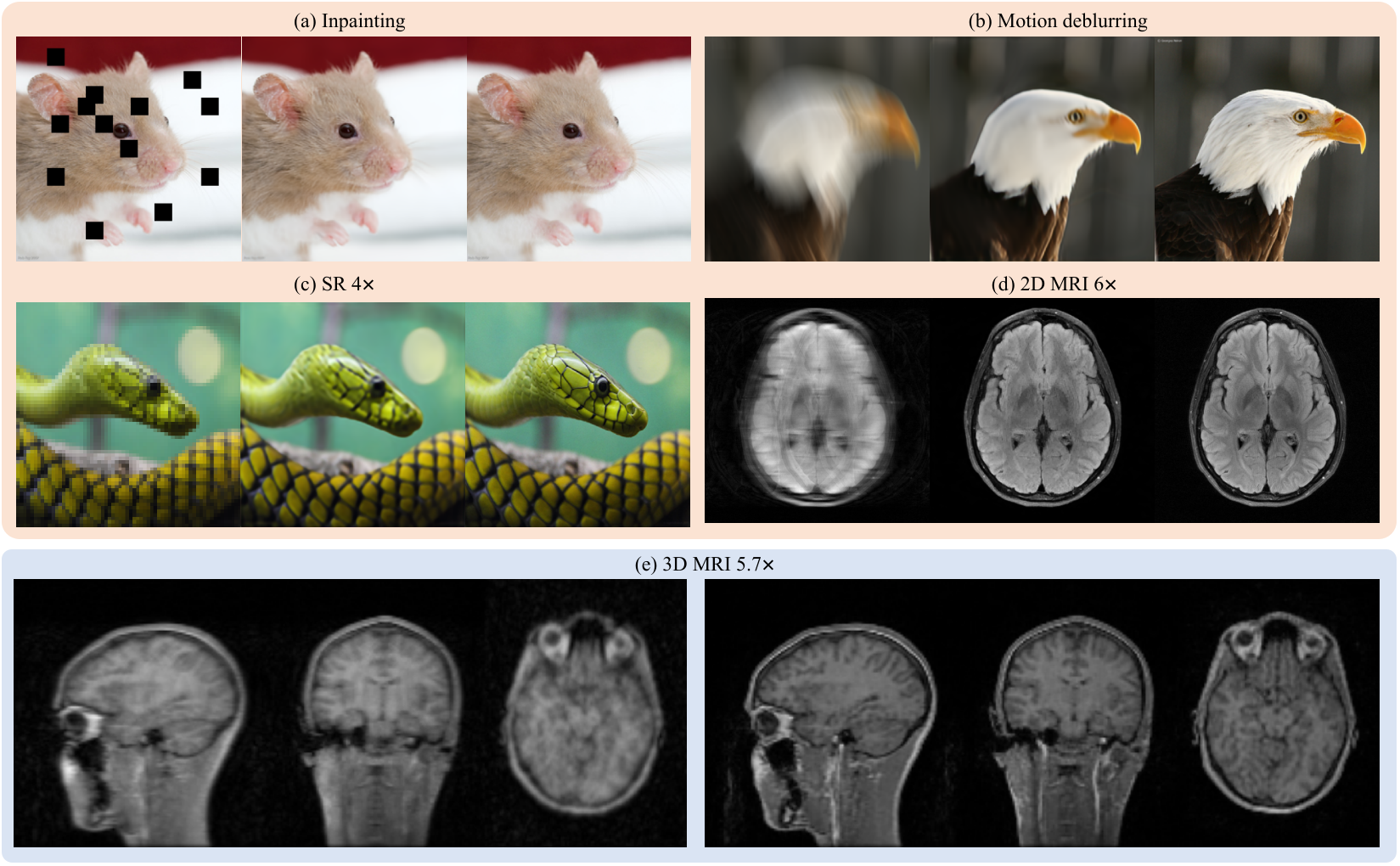}
    \vspace{-1em}
    \caption{ Demonstration of self-diffusion applied to various inverse problems across natural and medical imaging domains.
    (a) Image inpainting
    (b) Motion deblurring
    (c) 4× image super-resolution (SR)
    (d) 2D MRI reconstruction from 6$\times$ undersampled data
    (e) 3D MRI reconstruction from 5.7$\times$ undersampled data
    In each set, the leftmost image shows the degraded input, the middle shows the self-diffusion result, and the rightmost shows the ground-truth or reference.}
    \vspace{-1em}
\end{figure}
Recent advances in generative modeling have introduced a powerful alternative: using generative models as learned priors. Models such as variational autoencoders (VAEs) \citew{kingma2013auto}, generative adversarial networks (GANs) \citew{goodfellow2014generative}, and conditional autoregressive models \citew{van2016conditional} have shown impressive capabilities in capturing complex data distributions. In particular, diffusion models \citew{sohl2015dpm, ho2020denoising, song2021scorebased} have emerged as a leading class of generative models with abilities to produce high-quality samples. Leveraging the denoising step in diffusion models, recent works have adapted them for inverse problems \citew{luo2023bayesian, chung2023diffusion, rout2024solving}, using a pretrained model to guide the reconstruction process. These generative priors effectively encode structural knowledge of the data and yield robust reconstructions. However, their applicability depends heavily on access to large-scale curated datasets for training. When the data for training priors is limited or not available, these methods may not be suitable.

Interestingly, recent research has shown that deep neural networks can serve as strong implicit priors, even without pretraining. The Deep Image Prior (DIP) approach \citew{ulyanov2018deep, VanVeen2018, heckel2020compressive} exploits the tendency of overparameterized networks to capture natural image statistics. In DIP, a randomly initialized neural network is optimized directly on the observed measurements to minimize reconstruction error. Despite not being trained on external data, the network's inductive biases the solution toward natural-looking images. While promising, DIP's performance is sensitive to the optimization process and often suffers from premature convergence or limited expressiveness, which can lead to suboptimal results.

In this work, we propose self-diffusion, a novel approach for solving inverse problems without relying on pretrained generative models. Our method is motivated by the observation that diffusion models generate images progressively from low to high frequencies through the denoising process. Building on this insight, we design a self-diffusion process that iteratively reconstructs the source starting from noise and using a spectrum-regulated self-denoiser based on the DIP framework. During the process, a single neural network (serving as the denoiser) is continuously optimized to minimize the discrepancy between its estimated reconstruction and the observed data across different noise levels. This allows the model to refine the reconstruction in a progressive manner, guided entirely by the observed measurements, without any prior training or external datasets.
Our contributions are as follows:
\begin{enumerate}[leftmargin=1.5em,itemsep=0pt, topsep=0pt, parsep=0pt]
\item We introduce self-diffusion, a novel method for solving inverse problems that leverages the diffusion process without requiring pretrained generative models. To the best of our knowledge, this is the first such approach.
\item We provide a theoretical form of the self-diffusion process, offering a solid foundation for understanding its behavior and convergence properties from the perspective of spectral bias.
\item We evaluate self-diffusion on a range of inverse problems, including MRI reconstruction, low-level image restoration, and radar angle estimation. Experimental results demonstrate that our method achieves competitive performance compared to other methods.
\end{enumerate}

\begin{figure}[h]
    \centering
    \includegraphics[width=\textwidth]{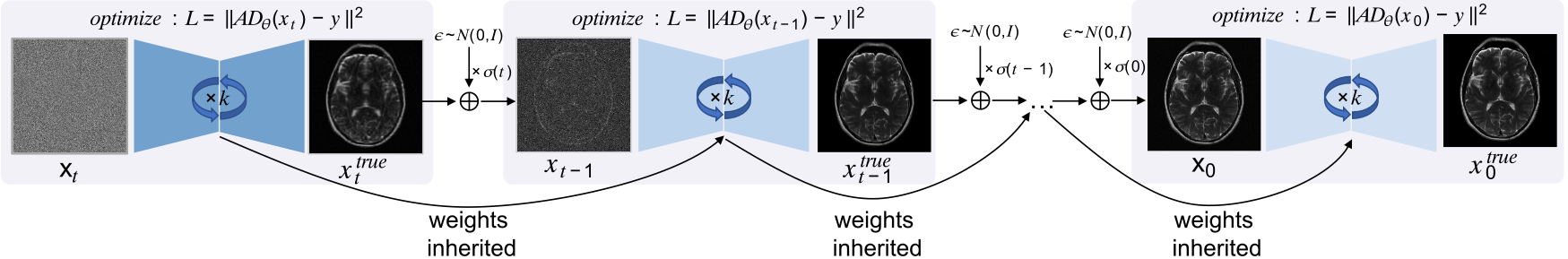}
    \vspace{-1.5em}
    \caption{Overview of the self-diffusion framework.
    At each diffusion step $t$, Gaussian noise is added to the current estimate $\mathbf{x}^{\text{true}}_t$, producing a noisy input. A single self-denoiser (initialized with an untrained network) is then trained to produce $\mathbf{x}^{\text{true}}_{t-1}$ by minimizing the task-specific data fidelity loss. This iterative alternation of noising and denoising from $\mathbf{x}_T$ to $\mathbf{x}_1$ yields a coarse-to-fine reconstruction process without the need for pretrained models or external supervision.
    }
\end{figure}

\section{Self-diffusion for inverse problem}
\label{sec:setup}
\paragraph{Problem Setup.} We consider the inverse problem defined by the equation
\[
    \mathcal{A}\mathbf{x}^{\text{true}} = \mathbf{y}~,
\]
where \(\mathcal{A} \in \mathbb{R}^{m \times n}\) is a known forward operator, \(\mathbf{x}^{\text{true}} \in \mathbb{R}^n\) is the unknown deterministic solution, and \(\mathbf{y} \in \mathbb{R}^m\) represents the observed data. In many practical scenarios, the problem is underdetermined, i.e., \(m < n\), which results in more unknowns than equations and makes the it potentially non-unique and ill-posed. To resolve this, regularization techniques are typically applied to introduce prior knowledge and ensure the existence and uniqueness of a stable solution.

Bayesian framework is adopted in many works, where we treat \(\mathbf{x}^{\text{true}}\) as a random variable with a prior distribution \(p(\mathbf{x})\), reflecting our prior knowledge about the solution. The observed data \(\mathbf{y}\) is modeled as an observation \(\mathcal{A}\mathbf{x}^{\text{true}}\), with likelihood function \(p(\mathbf{y} | \mathbf{x})\) that quantifies how probable the observed data is, given the solution \(\mathbf{x}\). Bayes' theorem allows us to compute the posterior distribution of \(\mathbf{x}^{\text{true}}\) given the observed data \(\mathbf{y}\) and the prior \(p(\mathbf{x})\)
\[
    p(\mathbf{x} | \mathbf{y}) = \frac{p(\mathbf{y} | \mathbf{x}) p(\mathbf{x})}{p(\mathbf{y})}~,
\]
where \(p(\mathbf{y}) = \int p(\mathbf{y} | \mathbf{x}) p(\mathbf{x}) \, d\mathbf{x}\) is the marginal likelihood. The solution \(\mathbf{x}^{\text{true}}\) is then obtained by maximizing the posterior distribution. In this framework, the prior $p(\mathbf{x})$ can be interpreted as a form of regularization, as it enforces certain desirable properties on the solution (such as smoothness, sparsity, or structure).
\paragraph{Self-diffusion.} 
To map the estimated noisy solution \( \mathbf{x}_t=\mathbf{x}^{\text{true}}_{t} + \sigma(t) \epsilon_t \) to the true solution \( \mathbf{x}^{\text{true}} \), a self-diffusion process trains a self-denoiser \( D_{\theta_{t,k}} \) at each noise step \( t = T-1, \ldots, 0 \) over \( k = 0, \ldots, K-1 \) iterations to minimize the loss
\begin{equation}
    L_{t,k} = \| \mathcal{A} D_{\theta_{t,k}}(\mathbf{x}^{\text{true}}_{t} + \sigma(t) \epsilon_t) - \mathbf{y} \|^2~,
    \label{eq:loss}
\end{equation}
where $ \mathbf{x}^{\text{true}}_{t} $ is the estimate of $ \mathbf{x}^{\text{true}} $ within noise step $t$, initialized as $ \mathbf{x}^{\text{true}}_{T} = \epsilon_0 $, with $ \epsilon_0 \sim \mathcal{N}(0, I) $. The noise 
$ \epsilon_t \sim \mathcal{N}(0, I) $ is sampled once at the start of noise timestep $ t $ and is fixed across all $ K $ iterations within that timestep. It is then resampled for the next noise step $ t-1 $. The noise schedule is \( \sigma_t = \sqrt{1 - \bar{\alpha}_t},~\text{where}~ \bar{\alpha}_t = \mathop{\textstyle\prod}\nolimits_{i=0}^t (1 - \beta_i)~\text{and}~\beta_t = \beta_{\text{end}} + \frac{t}{T - 1}(\beta_{\text{start}} - \beta_{\text{end}})\).
After $ K $ iterations at each noise step, the self-denoiser $ D_{\theta_{t,k}} $ produce the estimated solution $ \mathbf{x}^{\text{true}}_{t-1} $ for the next noise step $ t-1 $.
As the noise level $\sigma(t)$ decreases to zero, the predicted solution converges to the true solution $ \mathbf{x}^{\text{true}} $ under Theorem 1 in \citeauthor{heckel2020compressive} (\citeyear{heckel2020compressive}). The detailed demonstration of this process is in \cref{appendix:proof1}. The pseudo-code for implementation is shown in \Cref{alg:denoising}.

Since $\epsilon_t$ is fixed within each noise step but resampled at each subsequent noise step, the process $\mathbf{x}_t$ is piecewise stochastic. In the continuous limit, this process can be modelled with a stochastic differential equation (SDE) over time. Considering the noise resampling across steps, the SDE is
\begin{equation}
    d\mathbf{x}_t = 
    (\mathbf{x}_t - \mathbf{x}^{\text{true}}) dt + \sigma(t) dW(t)~,
\end{equation}
where $W(t)$ is a Wiener process capturing the resampling of noise across time steps. The drift term $(\mathbf{x}_t - \mathbf{x}^{\text{true}})$
reflects the self-denoiser's push toward $\mathbf{x}^{\text{true}}$. The derivation of this SDE is provided in \cref{appendix:proof2}.
This self-diffusion process is a combination of the forward and reverse processes because: 1) the forward process is self-referential ($\mathbf{x}_t = \mathbf{x}^{\text{true}}_{t} + \sigma(t) \epsilon_t$), not starting from the curated noise-free training dataset; 2) the proposed self-denoiser directly learns to predict $\mathbf{x}^{\text{true}}$ at each noise step instead of relying on the pretrained score function or denoiser.
\begin{algorithm}
    \caption{Self-diffusion (SDI) for solving the inverse problem}
    \label{alg:denoising}
    \begin{algorithmic}[1]
    \State \textbf{Input:} $\mathbf{x}^{\text{True}}_{T}$; noise steps - $T$; iterations-$K$; learning rate - $\eta$; forward operator - $\mathcal{A}$; initialize $\theta_{T,0}$; default noise schedule $\beta_{\text{start}} = 0.0001$, $\beta_{\text{end}} = 0.01$
    \For{$t = T-1$ to $0$}
    \State Sample $\epsilon_t \sim \mathcal{N}(0, I)$, $\mathbf{x}_{t} = (\mathbf{x}^{\text{true}}_{t} + \sigma(t) \epsilon_t)$ 
    \For{$k = 0$ to $K-1$}
        \State Compute loss $L_{t,k}$ and $\nabla_{\theta_{t,k}} L_{t,k}$
        \State Update the weights using optimizer and learning rate $\eta$
    \EndFor
    \State Set $\theta_t = \theta_{t,K}$
    \State Compute $\mathbf{x}^{\text{true}}_{t-1} = D_{\theta_{t}}(\mathbf{x}_{t})$
    \EndFor
    \State \Return $\mathbf{x}^{\text{true}}_{0}$
    \end{algorithmic}
    \end{algorithm}\vspace{-1em}
\paragraph{Noise-Modulated Spectral Bias.}
Neural networks, especially in overparameterized regimes, are known to exhibit spectral bias---a tendency to learn low-frequency components of a target function before high-frequency ones \cite{Soltanolkotabi2020Denoising}. This behavior has been theoretically linked to the eigenspectrum of the Neural Tangent Kernel (NTK) \citew{jacot2018neural}, which governs the dynamics of gradient descent during training. In particular, the NTK often has larger eigenvalues associated with smooth, low-frequency modes and smaller ones for high-frequency oscillations. Consequently, gradient updates are naturally biased toward reconstructing coarse, low-frequency structures earlier in training. 

The training of self-denoiser in self-diffusion exploits this spectral bias and further regulates it through the structured noise schedule. 
Considering the self-denoiser in \Cref{eq:loss}, we expand the expected loss function using a first-order Taylor expansion into
$$\mathbb{E}_{\epsilon_{t}}[||\mathcal{A}D_{\theta_{t,k}}(\mathbf{x}_{t})-\mathbf{y}||^{2}] \approx \underbrace{||\mathcal{A}D_{\theta_{t,k}}(\mathbf{x}_{t}^\text{true})-\mathbf{y}||^{2}}_{\text{Data Fidelity Term}} + \underbrace{\sigma(t)^{2} ||\mathcal{A}J_D(\mathbf{x}_{t}^\text{true})||_F^2}_{\text{Regularization Term}}$$
where the second term acts as a regularizer, discouraging the network from fitting sharp, high-frequency fluctuations. The derivation of this expansion is provided in \cref{appendix:proof3.1}.

This regularization effect is modulated by the noise level $\sigma_t$. In early steps of diffusion, where noise is large, the regularization is strong, enforcing smooth outputs and prioritizing low-frequency components. As the process progresses and $\sigma_t$ decreases, the regularization weakens, allowing the network to focus on finer, high-frequency details. This design leads to an implicit multi-scale learning regime, where the reconstruction transitions from global structure to local refinement. We reveal this process in Fourier space as detailed in \cref{appendix:proof3.2}.

\section{Simulation}
\label{sec:simulation}
We apply the self-diffusion inference to recover a sparse 1D signal from compressed sensing measurements. The original signal is generated as a sum of sine wave with varying frequencies and amplitudes and has a sparse representation in the frequency domain. 
The signal \( \mathbf{x} \) is constructed as
\[
\mathbf{x} = \sum_{(A, f) \in \mathcal{S}} A \cdot \sin\left(2\pi f \frac{t}{N} \right)~.
\]
We have a predefined set of amplitude-frequency pairs \( \mathcal{S} \) as
\[
\mathcal{S} = \left\{ 
(1.0, 1),\;
(0.5, 15),\;
(0.3, 20),\;
(1.0, 6),\;
(0.8, 3),\;
(0.6, 4),\;
(0.7, 5)
\right\}~,
\]
and \( N \) is the total signal length, with \( t \in \{0, 1, ..., N-1\} \). This signal is deliberately designed to span both low and high frequencies with non-uniform amplitudes to challenge the network's ability to recover both coarse and fine details from compressed measurements.
The signal is measured using a random Gaussian sampling matrix $\mathbf{A} \in \mathbb{R}^{m \times n}$, where $m \ll n$. The measurement matrix is normalized to have unit Frobenius norm. The measurements are then obtained as $\mathbf{y} = \mathbf{A} \mathbf{x}$. In this experiment, the signal length is 128, and the number of measurements is 35.

\begin{figure}[H]
    \centering
    \includegraphics[width=\textwidth]{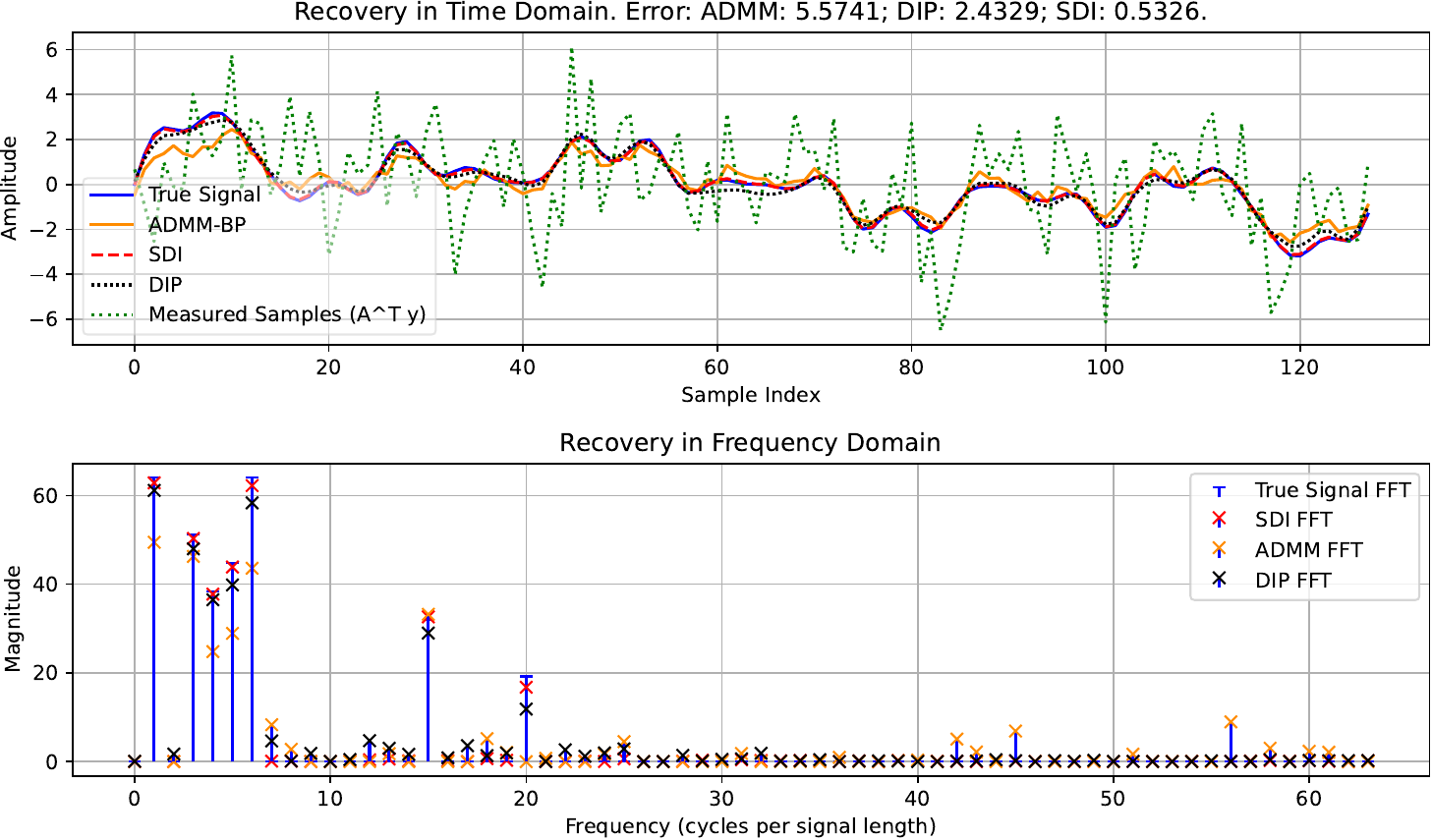}
    \vspace{-2pt}
    \caption{Recovery of a sparse signal composed of multiple sine waves from compressed measurements. The top plot illustrates time-domain reconstruction, comparing self-diffusion inference (SDI) to ADMM-basis pursuit (ADMM-BP) and Deep Image Prior (DIP). The bottom plot displays the frequency spectra, where SDI more accurately captures both low- and high-frequency components compared to ADMM-BP and DIP.\label{fig:1d_signal}}
\end{figure}

To recover the signal, we employ a 1D U-Net architecture with bottleneck layers and normalization layers as self-denoiser. All convolutional layers in the denoising network are initialized using a Gaussian distribution with zero mean and standard deviation of 0.02, while biases are initialized to zero. This initialization helps stabilize training and ensures the early network outputs maintain consistent scale and variance across layers. The parameters for \Cref{alg:denoising} is {\small$\{T=40; K=200; \eta=1e^{-5}; \beta_{\text{start}}=4e^{-3};\beta_{\text{end}}=1e^{-6}\}$} and Adam optimizer is used. We use DIP to reconstruct the signal with the same learning rate and initialized network as SDI. The loss used in both methods are enhanced with L1 penalty in frequency domain. We also run ADMM-basis pursuit in \citew{boyd2011distributed}, which is a classical compressed sensing method, as another baseline. \Cref{fig:1d_signal} shows the original signal and the recoverd signal in time domain and frequency domain. The progression of signal recovery over denoising steps is illustrated in the supplementary video \textit{freq\_prog.mp4}. In the early iterations, the model primarily recovers the low-frequency components of the signal, which dominate the coarse structure in the time domain. As the process continues, higher-frequency details gradually emerge. This progressive reconstruction---from low to high frequencies ---demonstrates the self-denoising capability of refining the initial estimate iteratively through successive denoising steps while benefitting from the spectral bias of the self-denoiser.
\section{Application}
\label{sec:application}
To demonstrate the versatility and generality of self-diffusion across a range of inverse problems spanning both natural and medical imaging domains, we evaluate our method on several representative tasks, including 2D and 3D MRI reconstruction, low-level vision restoration (inpainting, deblurring, denoising, super-resolution), and radar angle estimation recovery. The implementation details for each task are provided in the following sections. The code is available at \href{https://github.com/ggluo/self-diffusion}{github:ggluo/self-diffusion}.
\subsection{MRI reconstruction}
\paragraph{2D sampling.}
We evaluate the effectiveness of self-diffusion on 2D MRI reconstruction from undersampled k-space data. The raw k-space is subsampled using a equispaced Cartesian mask with an acceleration factor of 4x/6x along the phase-encoding direction. And 20 central auto-calibration signal (ACS) lines are obtained. From the fastMRI validation set \citew{zbontar2018fastmri}, we randomly selected 20 samples for each contrast (T1, T1 post contrast, T2, and FLAIR) to form a test set of 80 subjects. We estimated coil sensitivity maps estimated using BART's \texttt{ecalib} command. Measurements are formed with $\mathcal{A}$, which consists of the undersampling mask, Fourier transform, and coil sensitivities.

\begin{figure}
    \centering
    \includegraphics[width=0.75\textwidth]{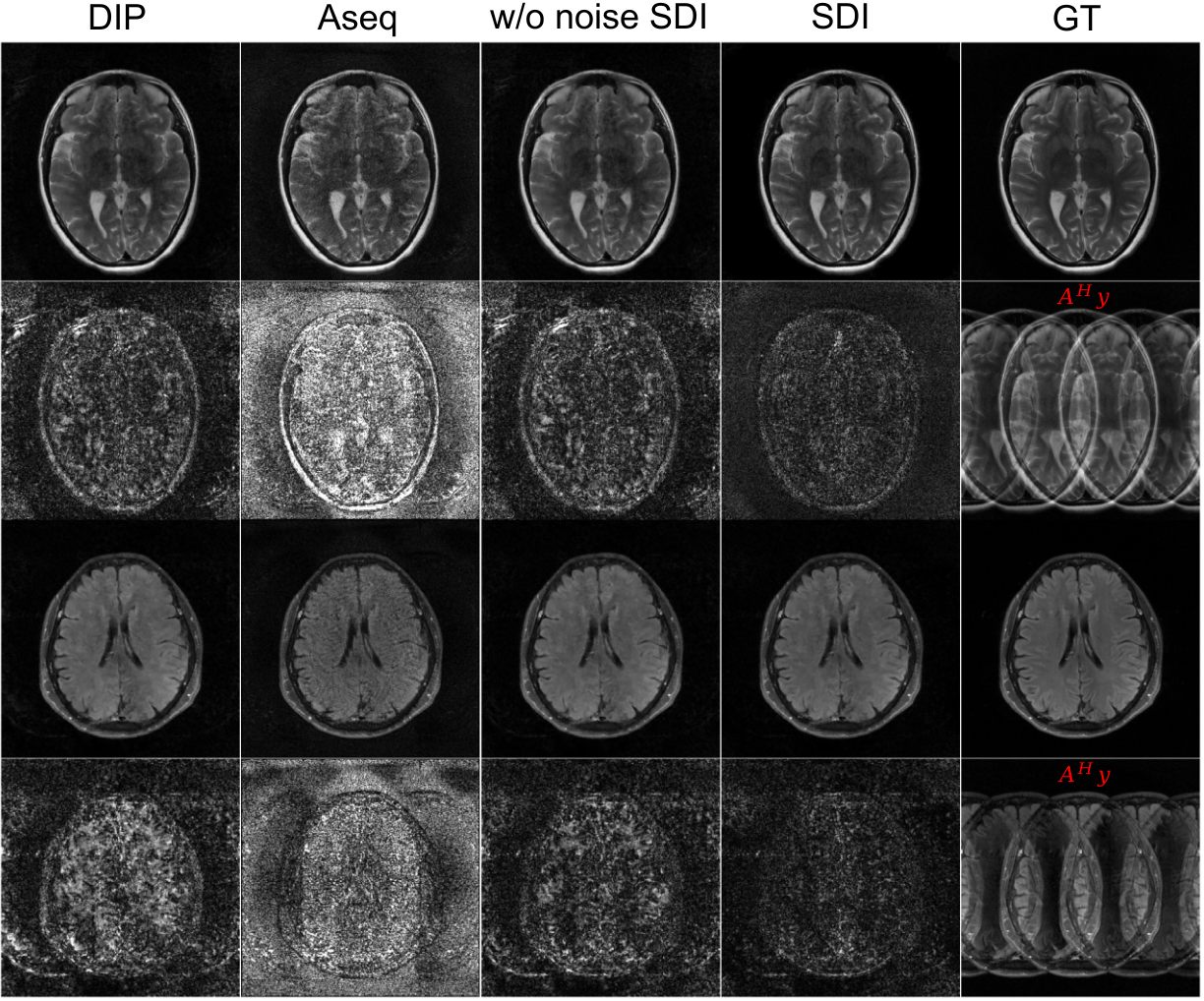}
    \caption{{Reconstructions and error maps to GT from 4$\times$ undersampled k-space w/o ACS lines. $\mathcal{A}^H \mathbf{y}$ is the degraded.}\label{fig:mri-4x-wo}}
\end{figure}

For self-diffusion, we initialize the 2D U-Net with Gaussian initialization \(\{\mu=0, \sigma=0.02\}\) for convolutional kernels and zero initialization for biases. 
To ensure stable optimization and consistent data fidelity weighting, we normalize the k-space measurements before input to the reconstruction process. 
Specifically, we compute a scale factor to match the norm of the initialized network's output to the norm of $\mathcal{A}^H\mathbf{y}$. The k-space measurements are scaled accordingly, preventing the network from unstable gradients due to mismatched signal energy.
We use Adam optimizer in \Cref{alg:denoising} and set parameters {\small$\{T=40; K=50; \eta=0.001; \beta_\text{start}=1e^{-3}; \beta_\text{end}=1e^{-4}\}$}. We compare SDI to the following baseline methods: Aseq \citew{alkhouri2024image}, IMJENSE \citew{Feng_TMI_2024}, and Deep Image Prior (DIP). For these Aseq and IMJENSE, we use authors's official implementations. The loss for DIP and SDI is enhanced with the total variation term, weight is set to 0.0001. 
As an ablation study, we also include SDI without resampling noise in \Cref{alg:denoising} and refer it as ``w/o $\epsilon_t$ SDI''.
The PSNR, SSIM, and NRMSE metrics are computed against the fully sampled ground truth.
\Cref{tab:mri} shows the reconstruction quality for all methods. SDI achieves the highest PSNR and SSIM, and the lowest NRMSE, outperforming Aseq and DIP that share the characteristic of using untrained networks but lack the iterative diffusion process.
\Cref{fig:mri-4x-wo} compares the reconstructed images for 2D MRI from 4$\times$ undersampled k-space data without ACS lines. Visually, SDI reconstructs sharper anatomical boundaries and finer textures with fewer aliasing artifacts compared to other baselines. In \Cref{sec:mri_reconstruction}, \Cref{fig:mri-4x-20,fig:mri-6x} illustrate the visual results for 4$\times$ and 6$\times$ undersampled k-space data, respectively, and \Cref{fig:evo} reveals the evolution of the estimate over the self-denoising process.
\vspace{-0.5em}
\begin{table}[H]
    \caption{{\small Comparison for MRI reconstruction from 4$\times$ and 6$\times$ undersampled data.}\label{tab:mri}}
    \centering
    \vspace{-6pt}
    \resizebox{0.68\textwidth}{!}{
    \begin{tabular}{c@{\hskip 10pt} c @{\hskip 10pt}c @{\hskip 10pt}c}%
        \toprule
           &  4$\times$ & 6$\times$ & 4$\times$ w/o ACS \\[0.5ex]
          Method &  PSNR↑/SSIM↑/NRMSE↓ & PSNR↑/SSIM↑/NRMSE↓ & PSNR↑/SSIM↑/NRMSE↓ \\\midrule
          $\mathcal{A}^H \mathbf{y}$ & 26.78 / 0.7411 / 0.0472 & 25.84 / 0.6990 / 0.0526 & 15.01 / 0.4546 / 0.1815 \\
          {\small IMJENSE} & 38.80 / 0.9604 / 0.0125 & 35.30 / 0.9398 / 0.0183 & N/A \\
          Aseq & 32.18 / 0.8661 / 0.0272 & 29.74 / 0.8249 / 0.0353 & 30.87 / 0.8460 / 0.0318  \\
          DIP & 38.35 / 0.9574 / 0.0136 & 34.92 / 0.9368 / 0.0192 & 36.34 / 0.9454 / 0.0178  \\
          {\small w/o $\epsilon_t$ SDI} & 39.09 / 0.9639 / 0.0120 & 36.21 / 0.9474 / 0.0165 & 37.12 / 0.9497 / 0.0160 \\
          SDI & 39.21 / 0.9640 / 0.0116 & 36.84 / 0.9492 / 0.0151 & 38.20 / 0.9541 / 0.0136\\
    \bottomrule 
    \end{tabular}}
    \vspace{-0.8em}
\end{table}
\paragraph{Comparison with methods using pretrained diffusion model.} We compared SDI with CSGM \citew{jalal2021robust} on reconstructing MR images of different contrast (T2 and FLAIR) as shown in \Cref{tab:con}. The diffusion model released in Ref. \citew{jalal2021robust} was trained on T2 contrast images. CSGM performs better in the reconstruction of T2 contrast images except in the case without ACS lines. CSGM requires 1150 the network forward evaluations per image. When CSGM apply this pre-trained model to FLAIR images, its performance is not as good as T2 contrast images. However, this does not mean that SDI is superior to methods using pretrained diffusion model as Ref \citew{luo2023bayesian} showed better cross-contrast generalization performance. Further domain-specific comparisons are omitted as they are beyond the scope of this work.
\begin{table}[H]
    \caption{{\footnotesize Compare SDI to CSGM on reconstructing different contrasted images. {\footnotesize (PSNR↑/SSIM↑/NRMSE↓)}}\label{tab:con}}
    \centering
    \vspace{-6pt}
    \resizebox{\textwidth}{!}{
    \begin{tabular}{c@{\hskip 10pt} c @{\hskip 10pt}c @{\hskip 10pt}c @{\hskip 10pt}c @{\hskip 10pt}c @{\hskip 10pt}c}%
        \toprule
          Acceleration & CSGM / T2 & {\small w/o $\epsilon_t$ SDI / T2} & SDI / T2 & {CSGM / FLAIR} & {\small w/o $\epsilon_t$ SDI / FLAIR} & SDI / FLAIR  \\[0.1ex]\midrule
          {4$\times$} & 38.78 / 0.9693 / 0.0119 & 38.30 / 0.9683 / 0.0135 & 38.08 / 0.9681 / 0.0131 & 37.41 / 0.9435 / 0.0153 & 37.95 / 0.9572 / 0.0138 & 38.41 / 0.9564 / 0.0131 \\
          {6$\times$} & 37.55 / 0.9636 / 0.0136 & 35.11 / 0.9519 / 0.0186 & 35.83 / 0.9555 / 0.0166 & 35.46 / 0.9286 / 0.0190 & 35.21 / 0.9367 / 0.0187 & 35.81 / 0.9387 / 0.0175 \\
          {4$\times$ w/o ACS} & 34.00 / 0.9376 / 0.0208 & 36.55 / 0.9543 / 0.0191 & 37.66 / 0.9620 / 0.0152 & 32.53 / 0.9015 / 0.0262 & 36.39 / 0.9454 / 0.0164 & 37.18 / 0.9408 / 0.0151 \\
    \bottomrule 
    \end{tabular}}
\end{table}

\subsection{Radar angle estimation recovery}
\begin{figure}[H]
    \centering
    \includegraphics[width=0.99\textwidth]{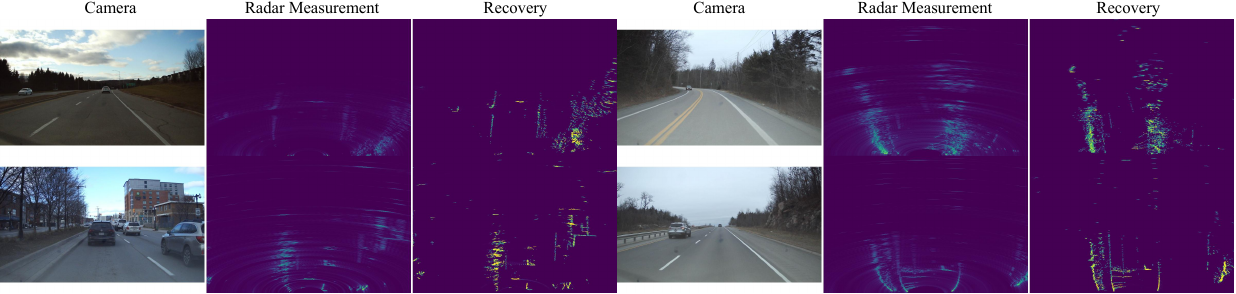}
    \vspace{-5pt}
    \caption{The camera column shows the visual context captured by the camera, depicting the road, vehicles, and surroundings. The radar measurement column displays the low-resolution radar range-azimuth heatmaps $\mathbf{y}$. The recovery column presents the recovered high-resolution angle estimations.}
    \label{fig:radar_recovery}
\end{figure}

Radar sensors are widely used in various applications, including automotive, aerospace, and etc. However, the resolution of radar angle estimation is often limited by the number of antennas (spatial sampling rate). This is a typical task that ground truth data is hard to obtain. Ref. \citew{yang2025unsupervisedradarpointcloud} formulates it as an inverse problem, and we apply self-diffusion to recover unknown high-resolution angle estimations $\mathbf{x}$ from low-resolution radar range-azimuth heatmaps $\mathbf{y}$ which represent the spatial locations of objects through signal power.
Following the setup in \citew{yang2025unsupervisedradarpointcloud}, we leverage a large-scale autonomous driving perception RADIal dataset \citew{rebut2022raw}. The measurement of low-resolution radar range-azimuth heatmap is synthesized by applying the radar angle measurement process to the objects that captured by LiDAR sensors under 86 antennas (spatial sampling rate). 
The same as previous settings, the denoising network is randomly initialized. The parameters for \Cref{alg:denoising} is {\small $\{T=30; K=100; \eta=1e^{-3}; \beta_{\text{start}}=2e^{-2};\beta_{\text{end}}=1e^{-4}\}$} and Adam optimizer is used. One thing deserves to be mentioned is that because the supposed output of our model is a 2D binary mask which can be easily thresholded to obtain the final point cloud, we use a sigmoid activation function as the last layer of the U-Net. \Cref{fig:radar_recovery} illustrates the radar recovery using self-diffusion on the RADIal dataset and four different scene from the dataset are shown. The results demonstrate significant resolution improvement despite hardware limitations making labeled data for radar enhancement challenging to obtain.

\subsection{Low-level vision tasks}
\begin{figure}[H]
    \centering
    \includegraphics[width=\textwidth]{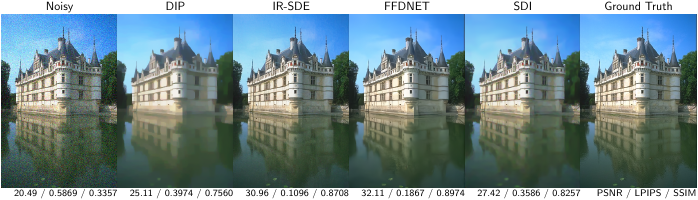}
    \vspace{-1.7em}
    \caption{Visual comparison of denoising results across different methods against the ground truth. IR-SDE by \citeay{luo2023image}; FFDNET by \citeay{zhang2018ffdnet}. \label{fig:denoise}}
\end{figure}

\paragraph{Evaluation.}
We evaluate self-diffusion on three representative low-level vision tasks: image inpainting, motion deblurring, and single-image super-resolution. Each task is formulated as $\mathbf{y} = \mathcal{A} \mathbf{x}$, where $\mathbf{x}$ is the unknown image, $\mathbf{y}$ is the observation and $\mathcal{A}$ is a task-specific degradation operator. The experiments are performed on 1000 images that are from the dataset ImageNet and resized to 256$\times$256.
For image inpainting, random rectangular masks remove regions of the image and the goal is to successfully fills missing regions with coherent structure and texture.
For motion deblurring, the blur is generated using code from \href{https://github.com/LeviBorodenko/motionblur}{Git:LeviBorodenko/motionblur} with kernel size 61 × 61 and intensity value 0.3, and goal is to reconstruct sharp images from blurry inputs.
In super-resolution tasks, low-resolution images are generated via average pooling and the goal is to recover high-resolution images from low-resolution inputs.
We use the \Cref{alg:denoising} for all the tasks above and set it with parameters {\small$\{T=40; K=150; \eta=0.001, \beta_{\text{start}}=1e^{-4}; \beta_{\text{end}}=1e^{-2}\}$}.
We also compare SDI to two baselines: DIP \citew{ulyanov2018deep} and Aseq \citew{alkhouri2024image}. The loss for DIP is enhanced with the total variation term. 
Evaluation metrics like PSNR, SSIM, and LPIPS are computed against the ground truth and are shown in \Cref{tab:low}. The visual results are illustrated in \Cref{fig:inpainting,fig:motion,fig:sr2,fig:sr4} listed in \Cref{sec:low_level_vision}.
\begin{table}[h]
\vspace{-0.5em}
\caption{Reconstruction quality comparison for low-level vision tasks. \small(PSNR↑ / SSIM↑ / LPIPS↓)\label{tab:low}}
\centering
\vspace{-5pt}
\resizebox{\textwidth}{!}{
\begin{tabular}{c@{\hskip 10pt} c@{\hskip 10pt}c@{\hskip 10pt}c@{\hskip 10pt}c@{\hskip 10pt}c}
\toprule
Task & $\mathbf{y}$ & Aseq & DIP & {\footnotesize w/o $\epsilon_t$ SDI} & SDI \\ [0.1ex]\midrule
2$\times$ SR  & 26.91 / 0.8924 / 0.0502 & 27.26 / 0.8359 / 0.0531 & 29.06 / 0.8530 / 0.0388 & 26.97 / 0.8852 / 0.0543 & 30.30 / 0.9186 / 0.0351 \\
4$\times$ SR  & 22.45 / 0.7392 / 0.0833 & 24.14 / 0.7535 / 0.0719 & 24.55 / 0.6855 / 0.0650 & 24.09 / 0.7583 / 0.0704 & 25.82 / 0.8110 / 0.0585 \\
Motion        & 18.84 / 0.5763 / 0.1272 & 22.65 / 0.6886 / 0.0840 & 25.94 / 0.7375 / 0.0589 & 26.92 / 0.8202 / 0.0513 & 27.76 / 0.8459 / 0.0481 \\
Inpainting    & 18.06 / 0.8845 / 0.1339 & 28.06 / 0.8589 / 0.0516 & 29.41 / 0.9310 / 0.0384 & 25.42 / 0.8840 / 0.0804 & 30.78 / 0.9296 / 0.0422 \\
\bottomrule
\end{tabular}
}
\vspace{-1.em}
\end{table}
\paragraph{Denoising.}
We included denoising results on CBSD68 dataset \cite{martin2001CB} with noise level $\sigma = 25$. SDI use the parameters {\small$\{T=30; K=100; \eta=0.002; \beta_\text{start}=1e^{-2}; \beta_\text{end}=8e^{-4} \}$}. DIP has $3000$ iterations enhanced with the total variation term whose weight is set to $5e^{-4}$. The other baselines use default parameters. As shown in \Cref{tab:denoising}, SDI performances worse than FFDNET and IR-SDE but better than DIP. The supervised method FFDNET provides the best performance in terms of PSNR and SSIM. IR-SDE is the best in terms of LPIPS. In addition to the numerical metrics, we provided visual comparisons in \Cref{fig:denoise}.
\vspace{-.5em}

\begin{wraptable}{r}{0.33\textwidth}
\vspace{-1.7em}
\caption{{\small Comparison for denoising\label{tab:denoising}}}
\centering
\resizebox{0.33\textwidth}{!}{
\begin{tabular}{c@{\hskip 7pt}c}
\toprule
Method & PSNR↑ / LPIPS↓ / SSIM↑ \\ [0.1ex]\midrule
DIP  &  25.44 / 0.307 / 0.6602 \\
FFDNET  &  31.22 / 0.121 / 0.8821 \\
IR-SDE  &  28.09 / 0.101 / 0.7866 \\
SDI    &  28.10 / 0.207 / 0.7860 \\
\bottomrule
\end{tabular}
}
\end{wraptable}

\begin{figure}
    \centering
    \includegraphics[width=\textwidth]{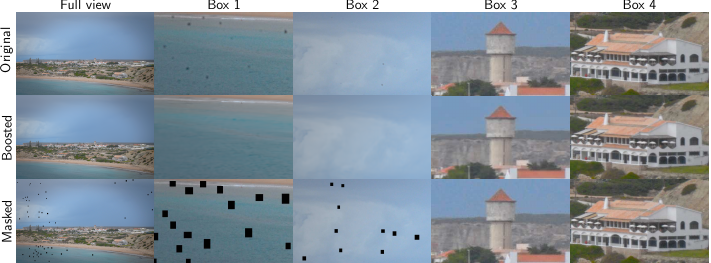}
    \vspace{-1.5em}
    \caption{SDI boosts image quality. The original image is corrupted by dust spots and minor noise. \label{fig:large}}
    \vspace{-1em}
\end{figure}

\paragraph{Large image}
We further evaluated SDI on a large image of size (5000$\times$3000), which is from an inexperienced shooter and corrupted by dust spots and minor noise. We masked the dust spots using a binary mask then restored the image using SDI via parameters {\small$\{T=40; K=500; \eta=0.001; \beta_\text{start}=1e^{-2}; \beta_\text{end}=1e^{-4} \}$}. Results are shown in \Cref{fig:large}. The noise on the cloudy sky and sea surface are removed, while the details on the architecture are preserved. The restoration takes around 2 hours on RTX A6000. As limited by the size of paper, we will post the boosted image at \href{{https://ggluo.github.io/projects/self-diffusion}}{ggluo:self-diffusion} for better display.
\subsection{Hyperparameters sensitivity}
\Cref{tab:sensitivity} presents PSNR (dB) values for various combinations of iterations (\( K \)) and noise steps (\( T \)) when reconstructing an MRI image from a 6x-undersampled k-space.
The following trends are observed: 1) For each fixed \( T \), PSNR generally increases with \( K \) up to a certain point, then plateaus or slightly declines;
2) Higher \( T \) values tend to yield higher PSNR at lower \( K \), but the advantage diminishes as \( K \) increases;
3) After reaching a peak, PSNR tends to stabilize or decrease slightly. Total iterations (\( T \times K \)) range from 250 (\( T=10, K=25 \)) to 20000 (\( T=40, K=500 \)). Generally, more iterations lead to higher PSNR before plateauing.
Therefore, when choosing hyperparameters, first select \( T \) to balance performance and computational cost, then consider \( K \) for optimal PSNR.

\begin{table}[h]
\centering
\caption{{\small PSNR for different iterations \( K \) and steps \( T \) choices.\label{tab:sensitivity}}}
\vspace{-0.5em}
\resizebox{0.7\textwidth}{!}{
\begin{tabular}{cccccccccc}
\toprule
\( T \backslash K \) & 25 & 50 & 75 & 100 & 150 & 200 & 300 & 400 & 500 \\
\midrule
10 & 24.65 & 28.89 & 31.16 & 33.12 & 34.41 & 35.11 & 35.95 & 35.96 & 36.06 \\
20 & 29.35 & 33.27 & 34.04 & 35.30 & 35.69 & 36.15 & 36.33 & 36.02 & 36.24 \\
40 & 32.13 & 34.54 & 35.60 & 35.96 & 36.19 & 36.17 & 36.07 & 36.04 & 36.11 \\
\bottomrule
\end{tabular}}
\end{table}

\Cref{tab:runtime} summarizes runtime and iteration comparison for different reconstruction methods.
\begin{table}[h]
\vspace{-0.5em}
\caption{{\small Runtime and iteration comparison for different reconstruction methods.\label{tab:runtime}}}
\centering
\vspace{-5pt}
\resizebox{\textwidth}{!}{
\begin{tabular}{c@{\hskip 10pt}c@{\hskip 10pt}c@{\hskip 10pt}c@{\hskip 10pt}c@{\hskip 10pt}c@{\hskip 10pt}c @{\hskip 10pt}c}
\toprule
 & IMJENSE & Aseq & CSGM & SDI & DPS & DDNM & SDI \\ [0.1ex] \midrule
Task, Size 
& MRI, 320$\times$320 
& MRI, 320$\times$320 
& MRI, 320$\times$320 
& MRI, 320$\times$320
& low-level vision, 256$\times$256 
& low-level vision, 256$\times$256
& inpainting, 5000$\times$3000 \\

Iter/sec 
& 11.92 
& 16.94 
& 11.39 
& 16.94
& 8.31 
& 19.64
& 1.52 \\

Total iters 
& 1500 
& 2000 
& 1156 
& 2000 
& 1000 
& 100 
& 20000 \\

GPU info 
& RTX 4090, 24GB 
& RTX 3090 Ti, 24GB 
& RTX 4090, 24GB 
& RTX 3090 Ti, 24GB 
& RTX 4090, 24GB 
& RTX 4090, 24GB 
& RTX A6000, 48GB \\
\bottomrule
\end{tabular}
}
\vspace{-1em}
\end{table}

\section{Discussions}
In this work, we proposed self-diffusion as a novel approach for solving inverse problems, which combines elements of denoising, diffusion processes, and the spectral bias of neural networks.
Self-diffusion leverages the spectral bias of overparameterized neural networks, which naturally prioritize low-frequency components during optimization. This bias is enhanced by a decreasing noise schedule, creating a hierarchical, coarse-to-fine reconstruction process. Early denoising steps focus on smooth, low-frequency structures, while later steps refine high-frequency details. This behavior mimics multiscale optimization---without requiring explicit architectural scale separation---and is well suited to many inverse problems as demonstrated by our experiments.

Compared to DIP, self-diffusion exhibits three key advantages:
1) Initialization robustness: DIP performance is sensitive to weight initialization, often requiring careful tuning. Self-diffusion, in contrast, is largely robust to random initializations due to its iterative, noise-modulated training scheme;
2) Frequency generalization: DIP tends to overfit high-frequency components early in training when data is noisy. Self-diffusion, through its structured noise schedule, maintains better control over frequency content during optimization;
3) Better zero-shot understanding: The noise in self-diffusion promotes understanding of image, which leads to more coherent structure and texture as shown in \Cref{fig:inpainting}.
While this is the opposite case for the "w/o noise SDI" and DIP.

The self-diffusion framework introduces a form of implicit regularization driven by noise. The denoiser's Jacobian is captured by the differential
$\nabla D_{\theta_{t,k}}$, and the term $\sigma(t)^{2} ||\mathcal{A}J_D(\mathbf{x}_{t}^\text{true})||_F^2$ quantifies the frequencies bias. 
This mechanismis naturally similar to the NTK perspective, where low-frequency modes have larger eigenvalues and thus converge faster. The regularization term amplifies this bias by suppressing first-order derivatives, effectively damping high-frequency components as shown in \Cref{appendix:proof3.2}.

While self-diffusion shows strong generalization and performance across different tasks, several promising directions remain for further exploration:\label{sec:limitations}
1) Adaptive noise scheduling:
learning task-specific or image-adaptive noise schedules may yield better convergence and fidelity;
2) Neural architecture search:
automatically discovering architectures better suited for self-diffusion could improve both efficiency and reconstruction quality;
3) Hybriding with pretrained models:
combining self-diffusion with pretrained features or priors may boost performance on more complex or semantic tasks.
\section{Conclusions}
Self-diffusion introduces a simple yet powerful framework for solving inverse problems without external data or pretrained models. By leveraging the intrinsic spectral bias of neural networks and regulating it through a structured noise schedule, it enables progressive, coarse-to-fine reconstruction. This approach achieves strong performance across diverse inverse problems like medical image reconstruction, signal recovery, and low-level vision tasks while remaining architecture- and data-agnostic. The results highlight the potential of self-supervised optimization guided by structured noise, opening new directions for learning-free and data-efficient image restoration.

\section*{Acknowledgements}
The rebuttal for this work's NeurIPS 2025 submission was done at the Luxembourg Institute of Health, with support from the National Research Fund of Luxembourg (FNR) under grant C24/IS/18942843.

\bibliography{references}
\appendix

\newpage
\section{Error Propagation under Ideal Denoiser Performance}
\label{appendix:proof1}
We follow the setup for self-diffusion described in \Cref{sec:setup}.
All solutions to the ill-posed inverse problem can be parameterized as:
\[ \mathbf{x} = \mathcal{A}^\dagger \mathbf{y} + \mathbf{z}_{\text{null}} \]
where \(\mathcal{A}^\dagger\) is the Moore-Penrose pseudoinverse of \(\mathcal{A}\), and \(\mathbf{z}_{\text{null}} \in \text{null}(\mathcal{A})\) is an arbitrary vector in the null space of \(\mathcal{A}\). The challenge is to identify \(\mathbf{x}^{\text{true}}\) from this affine solution space.
The self-diffusion algorithm iteratively refines an estimate of \(\mathbf{x}^{\text{true}}\). The process iterates backward in "noise time" steps \(t = T-1, T-2, \ldots, 0\). At each step \(t\), a neural network denoiser \(D_{\theta_{t,k}}\) (e.g., a CNN) is trained 
by minimizing the loss:
\[ L_{t,k} = \| \mathcal{A} D_{\theta_{t,k}}(\mathbf{x}^{\text{true}}_{t} + \sigma(t) \epsilon_t) - \mathbf{y} \|^2 \]
where \(\mathbf{x}^{\text{true}}_{t} \in \mathbb{R}^n\) is the image estimate at step \(t\). We initialize it  with \(\mathbf{x}^{\text{true}}_{T-1} = \epsilon_0, ~\epsilon_0 \sim \mathcal{N}(0, I_n)\) is a random noise vector.
\(\epsilon_t \sim \mathcal{N}(0, I_n)\) is a sample of Gaussian noise, fixed during the \(K\) inner training iterations within a given timestep \(t\), and resampled for the next step \(t-1\).
\(\sigma(t) > 0\) is a positive noise schedule, monotonically decreasing such that \(\sigma(t) \to 0\) as \(t \to 0\). 
\(\theta_{t,k}\) are parameters of the denoiser at inner iteration \(k\) of step \(t\). Then, the iterative update is given by:
\begin{enumerate}
    \item Sample \(\epsilon_t \sim \mathcal{N}(0, I_n)~\), set \(\mathbf{x}_{t} = \mathbf{x}^{\text{true}}_{t} + \sigma(t) \epsilon_t~\);
    \item For \(k = 0, \ldots, K-1\), compute the gradient of \(L_{t,k}\) w.r.t. \(\theta_{t,k}\)  and update parameters;
    \item Set the denoiser for step \(t\) as \(D_{\theta_t} = D_{\theta_{t,K}}~\);
    \item Update the image estimate via $\mathbf{x}^{\text{true}}_{t-1} = D_{\theta_t}(\mathbf{x}_{t})~.$
\end{enumerate}

We define the relative change error as 
$ \mathbf{e}_t = {\|\mathbf{x}^{\text{true}}_{t} - \mathbf{x}^{\text{true}}_{t-1}\|^2}/{\|\mathbf{x}^{\text{true}}_{t}\|^2}$.
In this section, we analyze the behavior of the relative change error \(\mathbf{e}_t\) under idealized denoiser performance 
i.e., the following holds as \(t \to 0\):
\begin{equation}
    \mathbb{E}_{\epsilon_t} [\mathbf{e}_t] \to 0 \quad \text{and} \quad \mathbb{E}_{\epsilon_t} [\| \mathbf{x}^{\text{true}}_{t} - \mathbf{x}^{\text{true}} \|^2] \to 0.
    \label{eq:convergence}
\end{equation}
\citeauthor{heckel2020compressive} (\citeyear{heckel2020compressive}) claimed that there exists a unique true image \(\mathbf{x}^{\text{true}}\) that satisfies \(\mathcal{A} \mathbf{x}^{\text{true}} = \mathbf{y}\) and untrained network possesses a structure (e.g., piecewise smoothness, low-frequency dominance) learnable by \(D_{\theta_t}\).
Therefore, we view the trained \(D_{\theta_t}\) act as a denoiser, implicitly favoring the structure of $\mathbf{x}^{\text{true}}$.
For each step \(t\), the \(K\) inner training iterations are sufficient for the denoiser \(D_{\theta_t}\) to approximately satisfy the measurement constraint \(\mathcal{A} D_{\theta_t}(\cdot) \approx \mathbf{y}\) when applied to inputs of the form \(\mathbf{x}^{\text{true}}_t + \sigma(t)\epsilon_t\).
The denoiser's error with respect to the true signal, $\delta_t(\mathbf{x}_{\text{noisy}}) := D_{\theta_t}(\mathbf{x}_{\text{noisy}}) - \mathbf{x}^{\text{true}}$, has an expected squared norm $\zeta_t := \mathbb{E}_{\epsilon_t} [\| \delta_t(\mathbf{x}^{\text{true}}_{t} + \sigma(t) \epsilon_t) \|^2]$.
We assume that $\zeta_t \to 0$ as $t \to 0$, when the measurements \(\mathbf{y} = \mathcal{A} \mathbf{x}\) is enough to recover the true signal using the implict structure of denoiser $D_{\theta_t}$.
In the following, we show that the convergence of \(\mathbf{e}_t\) in \Cref{eq:convergence} is determined by the denoiser's error with respect to the true signal, i.e., \(\zeta_t\).

\subsection{Update Rule and Error Dynamics}
Let \(\mathbf{d}_t = \mathbf{x}^{\text{true}}_{t} - \mathbf{x}^{\text{true}}\) be the absolute error of the estimate at step \(t\).
The update rule for the image estimate is:
\[ \mathbf{x}^{\text{true}}_{t-1} = D_{\theta_t}(\mathbf{x}^{\text{true}}_{t} + \sigma(t) \epsilon_t) \]
Using the definition of \(\delta_t\), we have:
\[ \mathbf{x}^{\text{true}}_{t-1} = \mathbf{x}^{\text{true}} + \delta_t(\mathbf{x}^{\text{true}}_{t} + \sigma(t) \epsilon_t)~. \]
Subtracting \(\mathbf{x}^{\text{true}}\) from both sides gives the error recurrence:
\[ \mathbf{d}_{t-1} = \delta_t(\mathbf{x}^{\text{true}}_{t} + \sigma(t) \epsilon_t) \]
Taking the expected squared Euclidean norm:
\[ \mathbb{E}_{\epsilon_t} [\|\mathbf{d}_{t-1}\|^2] = \mathbb{E}_{\epsilon_t} [\|\delta_t(\mathbf{x}^{\text{true}}_{t} + \sigma(t) \epsilon_t)\|^2] =\zeta_t \]
Shifting the index, we get the expected squared error for the iterate \(\mathbf{x}^{\text{true}}_t\):
\[ \mathbb{E}_{\epsilon_t} [\|\mathbf{d}_t\|^2] = \zeta_{t+1} \]

Now, let's analyze the relative change error \(e_t\)
\[ \mathbf{e}_t = \frac{\|\mathbf{x}^{\text{true}}_{t} - \mathbf{x}^{\text{true}}_{t-1}\|^2}{\|\mathbf{x}^{\text{true}}_{t}\|^2} = \frac{\|(\mathbf{x}^{\text{true}}_{t} - \mathbf{x}^{\text{true}}) - (\mathbf{x}^{\text{true}}_{t-1} - \mathbf{x}^{\text{true}})\|^2}{\|\mathbf{x}^{\text{true}}_{t}\|^2} = \frac{\|\mathbf{d}_t - \mathbf{d}_{t-1}\|^2}{\|\mathbf{x}^{\text{true}}_{t}\|^2}~. \]
The norms of the iterates remain bounded below by a positive constant, i.e., \(\|\mathbf{x}^{\text{true}}_{t}\|^2 \geq c^2 > 0\) for all relevant \(t\), which is reasonable as \(\mathbf{x}^{\text{true}}_{t} \to \mathbf{x}^{\text{true}}\) and \(\|\mathbf{x}^{\text{true}}\| > 0\). Then, we get
\[ \mathbb{E}_{\epsilon_t} [\mathbf{e}_t] \leq \frac{1}{c^2} \mathbb{E}_{\epsilon_t} [\|\mathbf{d}_t - \mathbf{d}_{t-1}\|^2] ~.\]
Using the inequality \(\|\mathbf{a} - \mathbf{b}\|^2 \leq 2\|\mathbf{a}\|^2 + 2\|\mathbf{b}\|^2\), we have
\[ \mathbb{E}_{\epsilon_t} [\mathbf{e}_t] \leq \frac{2}{c^2} (\mathbb{E}_{\epsilon_t} [\|\mathbf{d}_t\|^2] + \mathbb{E}_{\epsilon_t} [\|\mathbf{d}_{t-1}\|^2])~.\]
Substitute the expressions in terms of \(\zeta\):
\[ \mathbb{E}_{\epsilon_t} [\mathbf{e}_t] \leq \frac{2}{c^2} (\zeta_{t+1} + \zeta_t)~. \]
Since \(\zeta_t \to 0\) as \(t \to 0\), then 
\[ \mathbb{E}_{\epsilon_t} [\mathbf{e}_t] \to 0 \quad \text{as} \quad t \to 0 ~.\]
\subsection{Denoiser as a Proximal Operator}
With sufficient training iterations (i.e., large $K$), the denoiser learns to produce outputs that are consistent with the measurements.
This consistency enforced through the untrained network leads to implicitly favoring of smoothed images instead of the least-squares solution, which often suffers from artifacts and poor perceptual quality.
The optimization dynamics of training the denoiser impart a spectral bias on the solution, favoring smoother, low-frequency components. This behavior can be interpreted through the lens of proximal optimization, whereby the denoiser acts approximately as a proximal operator:
\[
D_{\theta_t}(\mathbf{x}_t) \approx \text{prox}_{\lambda_t R}(\mathbf{x}_t) = \arg \min_{\mathbf{z}} \frac{1}{2} \| \mathbf{z} - \mathbf{x}_t \|^2 + \lambda_t R(\mathbf{z}),
\]
where $R(\mathbf{z})$ is an implicit regularization functional that discourages undesirable features such as high-frequency noise or implausible structures. We will analyze in more detail, in \Cref{appendix:proof3}, how the spectral bias shaped by the noise schedule.

\section{Continuous Time Approximation}
\label{appendix:proof2}
The discrete update rule in previous section is written as
\begin{align*}
    \mathbf{x}^{\text{true}}_{t-1} -\mathbf{x}^{\text{true}}_{t} &= D_{\theta_t}(\mathbf{x}^{\text{true}}_{t} + \sigma(t) \epsilon_t) - \mathbf{x}^{\text{true}}_{t} ~.
    \end{align*}
Since \( t \) decreases from \( T \) to \( 0 \) and let $ x(t) = \mathbf{x}^{\text{true}}_{t} $, we have 
\[
x(t + \Delta t) - x(t) = x(t - 1) - x(t) = D_{\theta_t}(x(t) + \sigma(t) \epsilon_t) - x(t)~,
\]
where the change in \( t \) is negative (\( \Delta t = -1 \)).
In the continuous limit \( \Delta t \to 0 \), we have
\[
\frac{d x(t)}{dt} = \lim_{\Delta t \to 0} \frac{x(t + \Delta t) - x(t)}{\Delta t}~.
\]
Since \( \Delta t = -1 \), the right hand side becomes
\[
x(t - 1) - x(t) = D_{\theta_t}(x(t) + \sigma(t) \epsilon_t) - x(t)~,
\]
which leads to 
\[
\frac{d x(t)}{dt} \approx \frac{x(t) - x(t - 1)}{1} = -\left( D_{\theta_t}(x(t) + \sigma(t) \epsilon_t) - x(t) \right)~.
\]
With $D_{\theta_t}(x(t) + \sigma(t) \epsilon_t) = \mathbf{x}^{\text{true}} + \delta_t(x(t) + \sigma(t) \epsilon_t)$, we have
\begin{align*}
\frac{d x(t)}{dt} &= -\left( (\mathbf{x}^{\text{true}} + \delta_t(x(t) + \sigma(t) \epsilon_t)) - x(t) \right)\\
&= x(t) - \mathbf{x}^{\text{true}} - \delta_t(x(t) + \sigma(t) \epsilon_t)
\end{align*}
As \( t \to 0 \), \( \sigma(t) \to 0 \), and \( \delta_t \to 0 \), we have the ordinary differential equation 
\[
\frac{d x(t)}{dt} \approx x(t) - \mathbf{x}^{\text{true}}~.
\]
Since \( \epsilon_t \) is fixed within each noise step but resampled at each new noise step, the process is piecewise stochastic. In the continuous limit, we model the resampling as a stochastic process over time. The SDE form, considering the noise resampling across steps, is
\[
d x(t) = (x(t) - \mathbf{x}^{\text{true}}) dt + \sigma(t) dW(t)
\]
where \( W(t) \) is a Wiener process capturing the resampling of noise across noise steps. The drift term \( x(t) - \mathbf{x}^{\text{true}} \) reflects the denoiser's push towards \( \mathbf{x}^{\text{true}} \), and the diffusion term \( \sigma(t) dW(t) \) models the noise resampling.

\section{Derivation of Noise-Modulated Spectral Bias}
\label{appendix:proof3}
The regularization term,as described in the previous section, arises from the noise added to the input of the self-denoiser and plays a critical role in enhancing the spectral bias. The term is derived from the expected loss over the noise distribution and is approximated as a regularizer. Below, we elaborate on the regularization term, its mathematical form, its effect in the Fourier domain, and its role in the self-diffusion process.

\subsection{Regularization Perspective}
\label{appendix:proof3.1}
To find a second-order approximation for the expected loss, $\mathbb{E}_{\epsilon_{t}}[||\mathcal{A}D(x_{t})-y||^{2}]$, where $\mathbf{x}_t = \mathbf{x}_{t}^\text{true} + \sigma(t)\epsilon_{t}$ and $\epsilon_t \sim \mathcal{N}(0, I)$. For clarity, let's simplify the notation for a single step $t$:
$D(\cdot) = D_{\theta_{t,k}}(\cdot)$, $x_0 = \mathbf{x}_{t}^\text{true}$, $\sigma = \sigma(t)$, $\epsilon = \epsilon_t$, and let the full transformation be $f(x) = \mathcal{A}D(x)$. The loss is $L = ||f(x_0 + \sigma\epsilon) - y||^2$. We want to compute $\mathbb{E}_{\epsilon}[L]$.

We start by approximating the output of the function $f(x)$ for a small perturbation $\sigma\epsilon$ around the point $x_0$. Using a first-order multivariate Taylor expansion, we get:
$$f(x_0 + \sigma\epsilon) \approx f(x_0) + J_f(x_0) (\sigma\epsilon)$$
where $J_f(x_0)$ is the Jacobian matrix of the function $f$ evaluated at $x_0$. The $(i, j)$-th element of $J_f$ is $\frac{\partial f_i}{\partial x_j}$. Now, we substitute this approximation back into the loss expression:
$$L \approx || (f(x_0) - y) + \sigma J_f(x_0) \epsilon ||^2~.$$
To simplify, let's define the residual vector $r = f(x_0) - y$. The expression becomes:
$$L \approx || r + \sigma J_f(x_0) \epsilon ||^2~.$$
We can expand the squared Euclidean norm (dot product) as $(v)^T(v)$:
$$L \approx (r + \sigma J_f(x_0) \epsilon)^T (r + \sigma J_f(x_0) \epsilon)~,$$
$$L \approx r^T r + 2\sigma r^T J_f(x_0) \epsilon + \sigma^2 (J_f(x_0) \epsilon)^T (J_f(x_0) \epsilon)~,$$
$$L \approx ||r||^2 + 2\sigma r^T J_f(x_0) \epsilon + \sigma^2 \epsilon^T J_f(x_0)^T J_f(x_0) \epsilon~.$$
This gives us three terms to analyze. We now take the expectation of the loss $L$ with respect to the noise distribution $\epsilon \sim \mathcal{N}(0, I)$.

1. The residual $r = f(x_0) - y$ does not depend on $\epsilon$.
    $$\mathbb{E}_{\epsilon}[||r||^2] = ||r||^2 = ||f(x_0) - y||^2$$

2. This term is linear in $\epsilon$. Since the expectation of the noise is zero, $\mathbb{E}_{\epsilon}[\epsilon]=0$, this term vanishes.
    $$\mathbb{E}_{\epsilon}[2\sigma r^T J_f(x_0) \epsilon] = 2\sigma r^T J_f(x_0) \mathbb{E}_{\epsilon}[\epsilon] = 0$$

3.  This term is a quadratic form in $\epsilon$. For a random vector $\epsilon \sim \mathcal{N}(0, I)$ and a constant matrix $Q$, we have the property $\mathbb{E}[\epsilon^T Q \epsilon] = \text{Tr}(Q)$.
    $$\mathbb{E}_{\epsilon}[\sigma^2 \epsilon^T J_f(x_0)^T J_f(x_0) \epsilon] = \sigma^2 \text{Tr}(J_f(x_0)^T J_f(x_0))$$
The trace of $J^T J$ is equal to the squared Frobenius norm of the matrix $J$, denoted $||J||_F^2$ by convention.
$$\sigma^2 \text{Tr}(J_f(x_0)^T J_f(x_0)) = \sigma^2 ||J_f(x_0)||_F^2$$

Combining the expectations of the three terms, we arrive at the final approximation for the expected loss:
$$\mathbb{E}_{\epsilon}[L] \approx ||f(x_0) - y||^2 + \sigma^2 ||J_f(x_0)||_F^2$$
Substituting back our original notation ($f(x) = \mathcal{A}D(x)$, $\mathbf{x}_t = \mathbf{x}_{t}^{true} + \mathbf{\epsilon}_t$ and $J_f = J_{\mathcal{A}D} = \mathcal{A}J_D$):
$$\mathbb{E}_{\epsilon_{t}}[||\mathcal{A}D(\mathbf{x}_{t})-y||^{2}] \approx \underbrace{||\mathcal{A}D(\mathbf{x}_{t}^\text{true})-y||^{2}}_{\text{Data Fidelity Term}} + \underbrace{\sigma(t)^{2} ||\mathcal{A}J_D(\mathbf{x}_{t}^\text{true})||_F^2}_{\text{Regularization Term}}$$

This term penalizes the squared Frobenius norm of the Jacobian of the entire transformation from the denoiser's input to its final output.
The Jacobian measures how much the output changes in response to small changes in the input. Penalizing its norm forces the learned function $D$ to be smoother and less sensitive to high-frequency noise.
This directly achieves the goal of modulating the spectral bias. By penalizing large derivatives, the optimization is biased toward learning low-frequency components first, especially when the noise level $\sigma(t)$ is high. This provides a rigorous foundation for the coarse-to-fine reconstruction behavior described in the previous section.

\subsection{Fourier Domain Analysis}
\label{appendix:proof3.2}

To understand the effect of the regularization term, we decompose the denoiser output in the orthonormal Fourier basis $\{ \mathbf{e}_k \}_{k=1}^n$, where each $\mathbf{e}_k$ corresponds to a Fourier mode with frequency index $k$. The denoiser's output can then be expressed as
\[
D(\cdot) = \sum_{k=1}^{n} d_{t,k} \, \mathbf{e}_{k},
\]
where $d_{t,k}$ are the Fourier coefficients at time $t$. The derived regularizer is
\[
R(D) = \sigma(t)^{2} \, \| \mathcal{A} J_D(\mathbf{x}_{t}^{\text{true}}) \|_F^2,
\]
where $J_D$ denotes the Jacobian of the denoiser $D$, $\mathcal{A}$ is a known linear operator, and $\|\cdot\|_F$ is the Frobenius norm.
A first-order differential operator acts on a Fourier mode $\mathbf{e}_k \sim e^{i k \cdot x}$ by scaling its amplitude by its frequency magnitude $|k|$. Hence, by Parseval's theorem, the squared $L_2$-norm of the gradient of $D$ can be expressed in terms of its Fourier coefficients as
\[
\| \nabla D \|^2 = \int \| \nabla D(x) \|^2 \, dx \propto \sum_{k} |k|^2 |d_{t,k}|^2.
\]
Since the Frobenius norm of the Jacobian, $\| J_D \|_F^2$, equals the sum of the squared $L_2$-norms of the gradients of each output component, it follows that
\[
\| J_D \|_F^2 \propto \sum_{k} |k|^2 |d_{t,k}|^2.
\]
Thus, the Jacobian term introduces a frequency-dependent quadratic penalty on the Fourier coefficients, scaling as $k^2$.
Substituting this into the regularizer gives
\[
R(D) = \sigma(t)^2 \, \| \mathcal{A} J_D \|_F^2 
\;\;\Rightarrow\;\;
R(D) \propto \sigma(t)^2 \sum_{k} |k|^2 \, \| \mathcal{A}\mathbf{e}_k \|^2 |d_{t,k}|^2,
\]
showing that higher-frequency modes are penalized more strongly. Moreover, the penalty strength decreases over time through the factor $\sigma(t)^2$.
We can therefore write an effective loss in the Fourier domain as:
\[
\mathbb{E}_{\epsilon_t}[L_{t,k}] 
\approx 
\underbrace{\sum_{k=1}^{n} \| \mathcal{A}\mathbf{e}_{k} \|^{2} (d_{t,k} - c_k)^{2}}_{\text{Data Fidelity}}
+
\underbrace{\sigma(t)^{2} \sum_{k=1}^{n} \kappa |k|^{2} \| \mathcal{A}\mathbf{e}_{k} \|^{2} |d_{t,k}|^{2}}_{\text{Jacobian Regularizer}},
\]
where $c_k$ are the Fourier coefficients of the true signal and $\kappa$ is a proportionality constant. This formulation confirms that the noise-modulated regularizer naturally suppresses high-frequency components, with a time-dependent weight controlled by $\sigma(t)^2$.

\section{Evolution of the reconstruction process}
\begin{figure}[H]
    \centering
    \includegraphics[width=\textwidth]{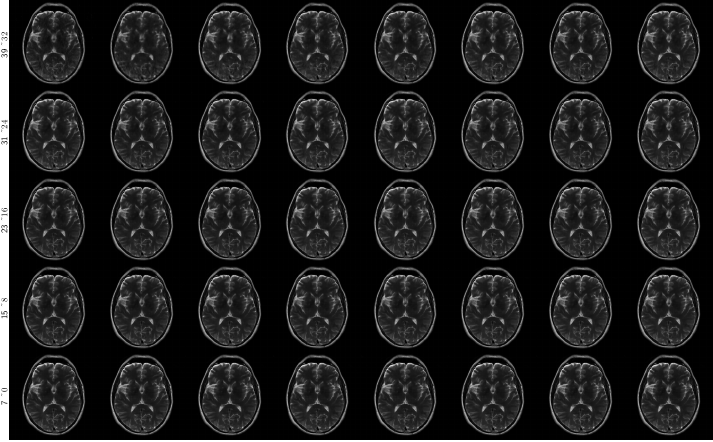}
    \caption{The evolution of reconstruction $\mathbf{x}_t^{\text{true}}$ over noise timestep from 39 to 0.\label{fig:evo}}
\end{figure}
\vspace{6cm}

\section{MRI Reconstruction}
\label{sec:mri_reconstruction}
\begin{figure}[H]
    \centering
    \includegraphics[width=0.85\textwidth]{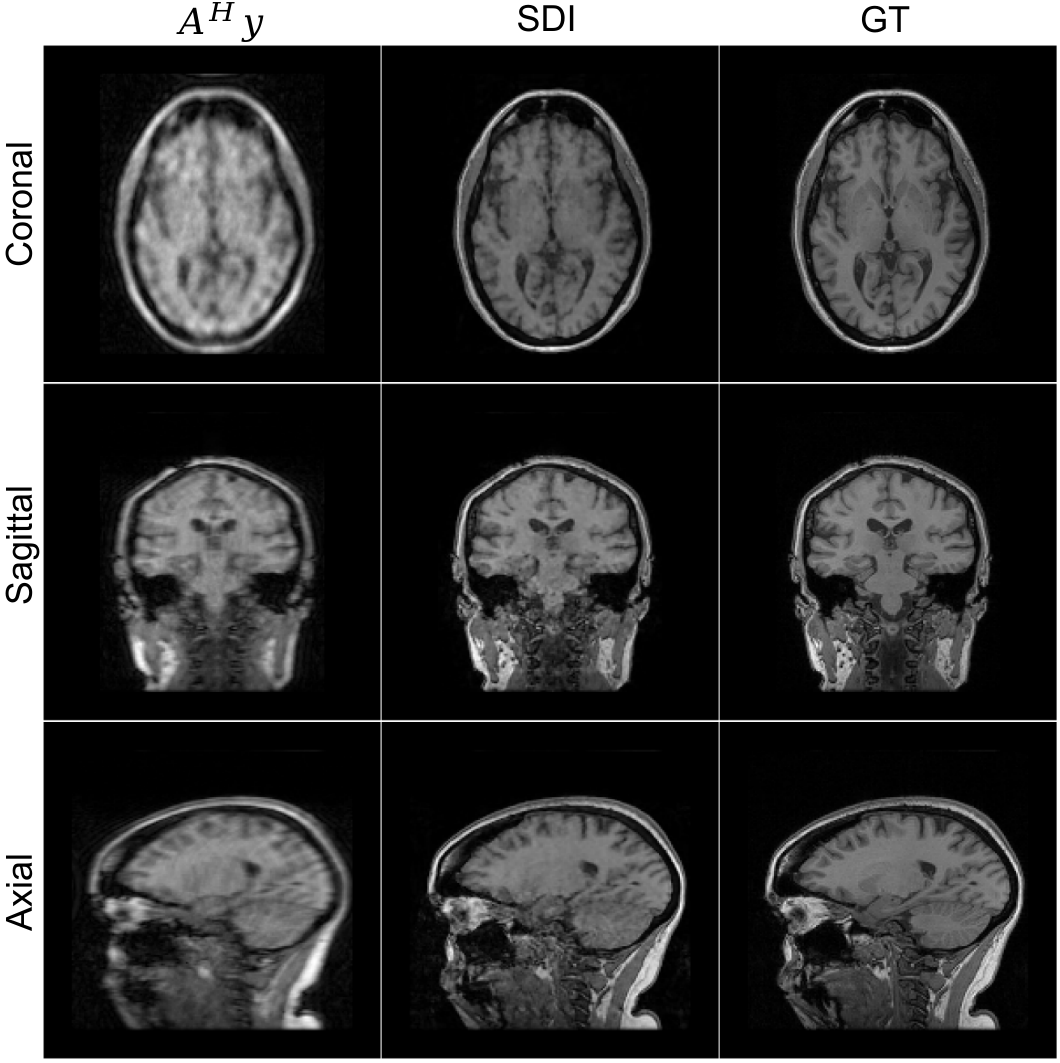}
    \vspace{-.5em}
    \caption{Reconstruction of a 3D T1-weighted brain MRI using self-diffusion from 8$\times$ undersampled measurements. Shown are sagittal, coronal, and axial slices, demonstrating sharp anatomical structures and minimal artifacts. The initial measurement ($\mathcal{A}^H\mathbf{y}$), self-diffusion inference (SDI), and ground truth (GT).}
    \vspace{-1.5em}
    \label{fig:3d_mri}
\end{figure}

\paragraph{3D sampling.}
To further demonstrate the flexibility of self-diffusion, we apply it to the reconstruction of volumetric 3D MRI data from undersampled k-space measurements. This task poses additional challenges due to the increased dimensionality and the need for spatial coherence across slices. We consider a 12-coil 3D T1-weighted brain scan and apply a Cartesian undersampling mask on the k-space that has a dimension of 160$\times$160$\times$128. The undersampling mask is generated using BART's command \texttt{poisson} with an acceleration factor of 2.5 along two phase-encoding directions. This leads to a total acceleration factor of 8. The coil sensitivities are estimated using BART's command \texttt{ecalib}. The measurements $\mathbf{y}$ are formed with $\mathcal{A}$, which consists of the undersampling mask, 3D Fourier transform, and coil sensitivities.
To accommodate 3D dimensionality, we adapt the denoising network to a 3D U-Net architecture with 3D convolutional layers and which is initialized in the same way as the 2D U-Net.
The parameters for \Cref{alg:denoising} are {\small$\{T=40; K=50; \eta=0.001\}$} and Adam optimizer is used.
Figure~\ref{fig:3d_mri} shows the reconstructed sagittal, coronal, and axial slices. 

\begin{figure}[H]
    \centering
    \includegraphics[width=0.8\textwidth]{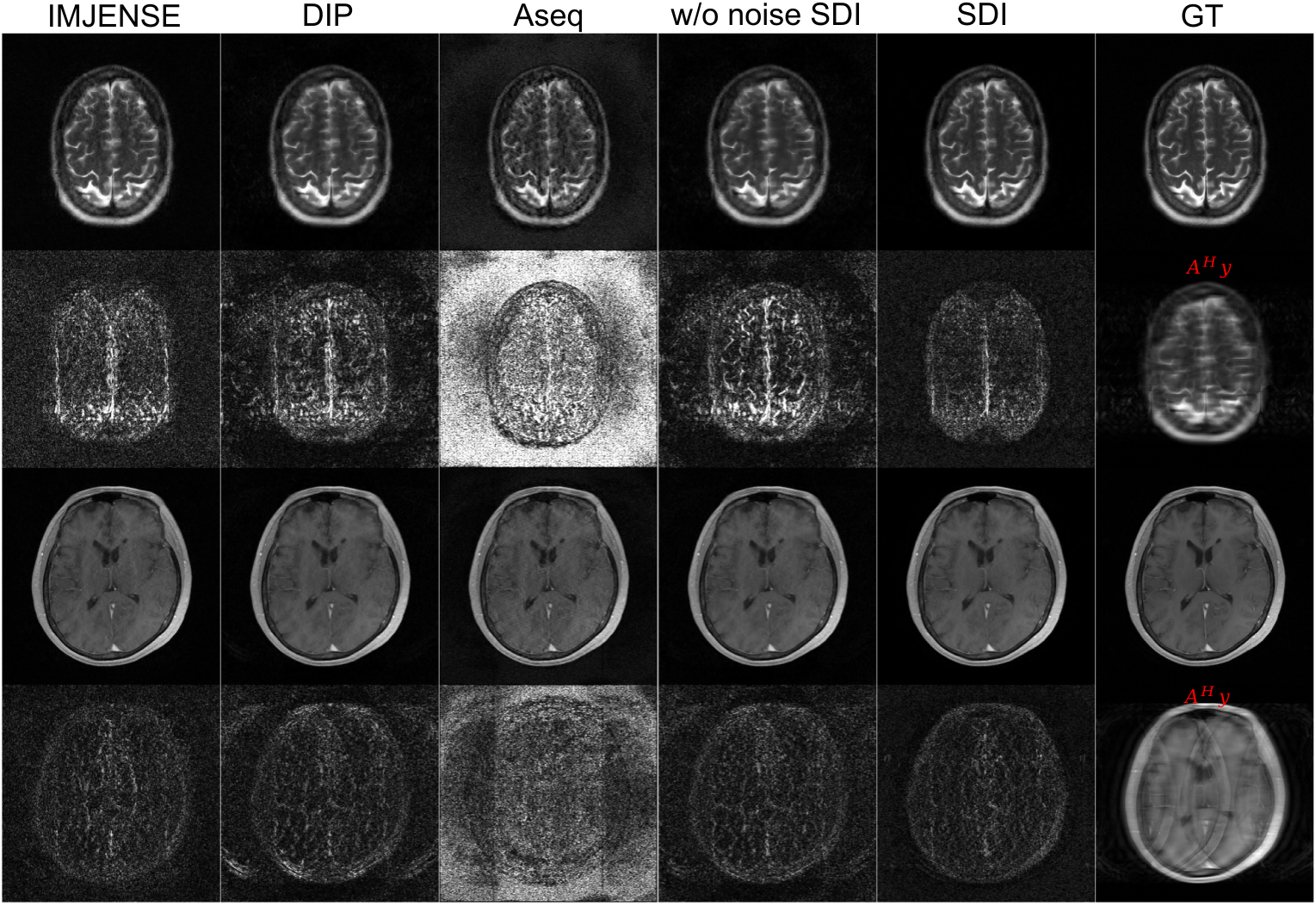}
    \caption{Reconstruction from 4$\times$ undersampled k-space with 20 ACS lines using different methods ($\mathcal{A}^H\mathbf{y}$,IMJENSE, DIP, Aseq, w/o noise SDI, SDI) and corresponding error maps are shown.\label{fig:mri-4x-20}}
\end{figure}
\begin{figure}[H]
    \centering
    \includegraphics[width=\textwidth]{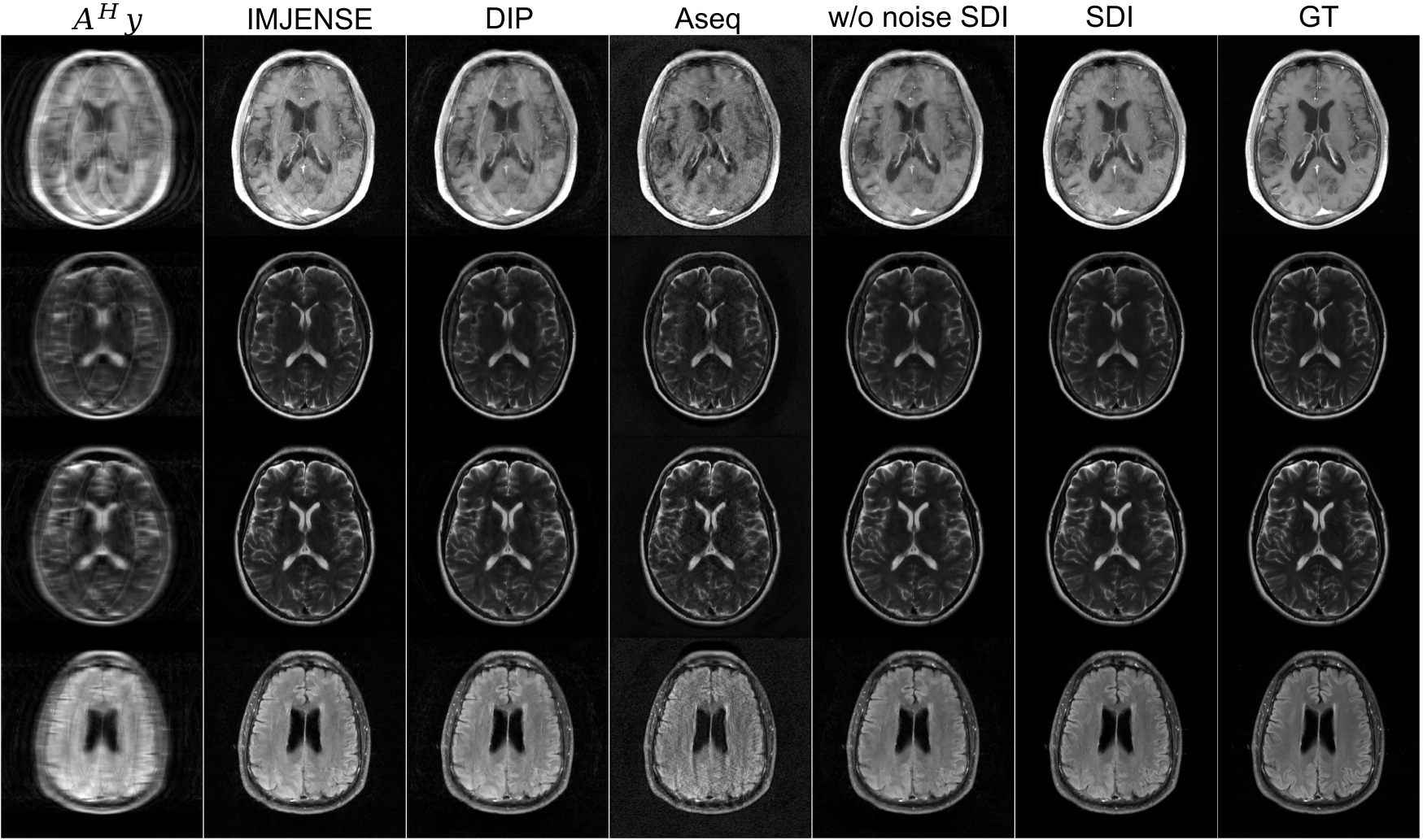}
    \caption{Reconstruction from 6$\times$ undersampled k-space with 20 ACS lines using different methods (IMJENSE, DIP, Aseq, w/o noise SDI, SDI).\label{fig:mri-6x}}
\end{figure}
\section{Low-level Vision Tasks}
\label{sec:low_level_vision}
\begin{figure}[H]
    \centering
    \includegraphics[width=\textwidth]{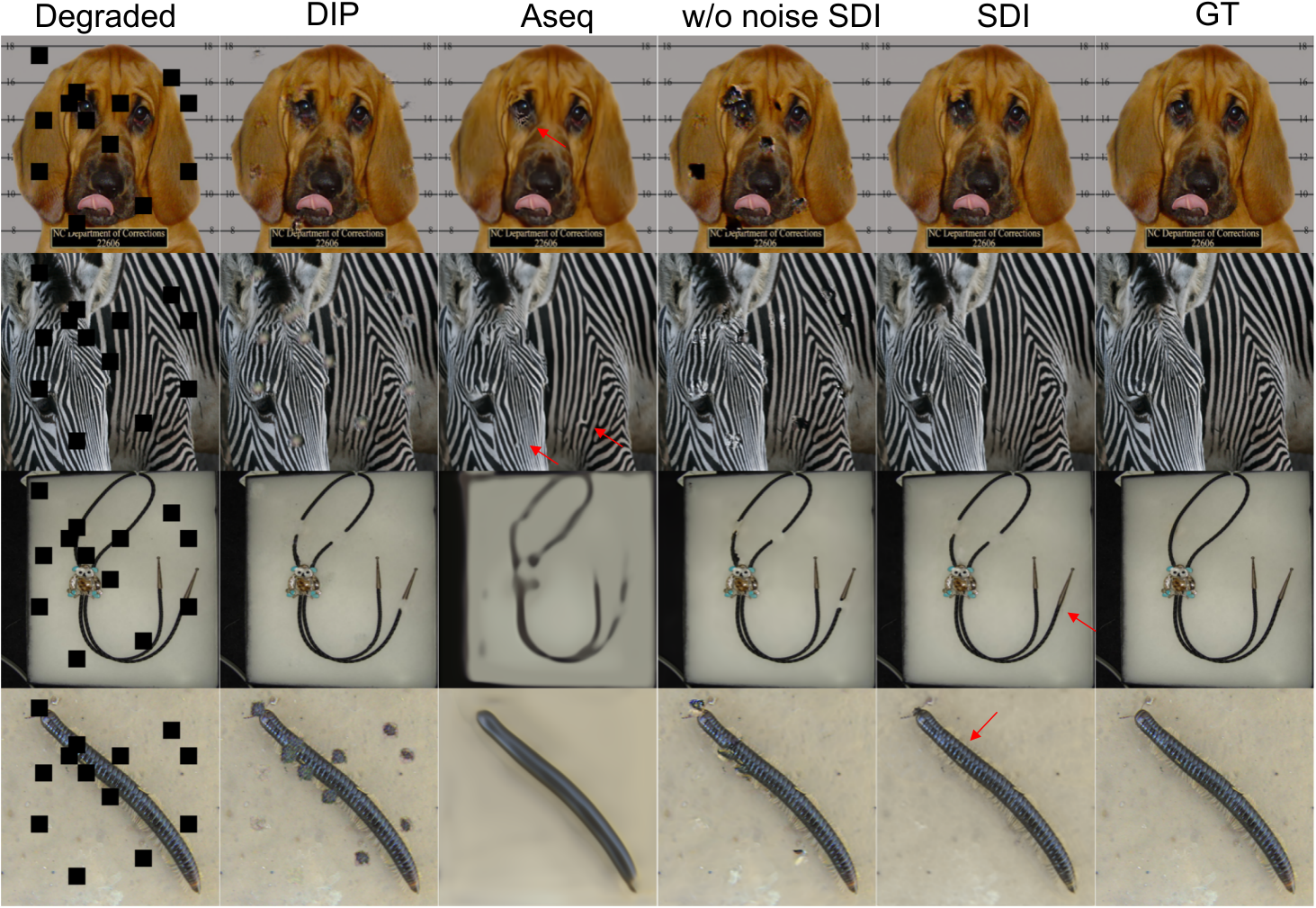}
    \caption{Comparison of inpainting results across different methods (DIP, Aseq, w/o noise SDI, SDI) against the ground truth (GT).\label{fig:inpainting}}
  \end{figure}
\begin{figure}[H]
    \centering
    \includegraphics[width=\textwidth]{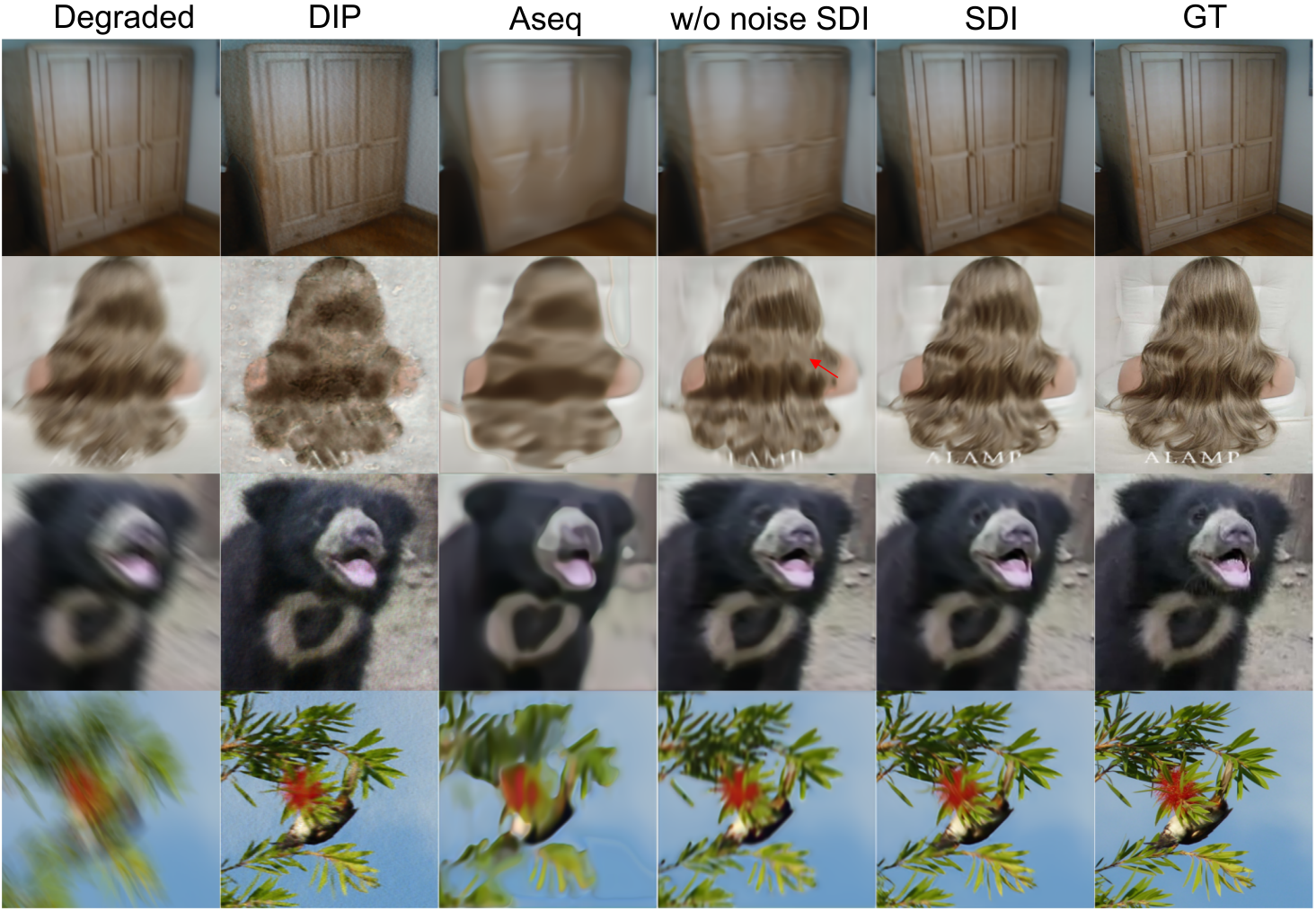}
    \vspace{-1em}
    \caption{Comparison of motion deblurring results across different methods (DIP, Aseq, w/o noise SDI, SDI) against the ground truth (GT).\label{fig:motion}}
  \end{figure}
\begin{figure}[H]
    \centering
    \includegraphics[width=\textwidth]{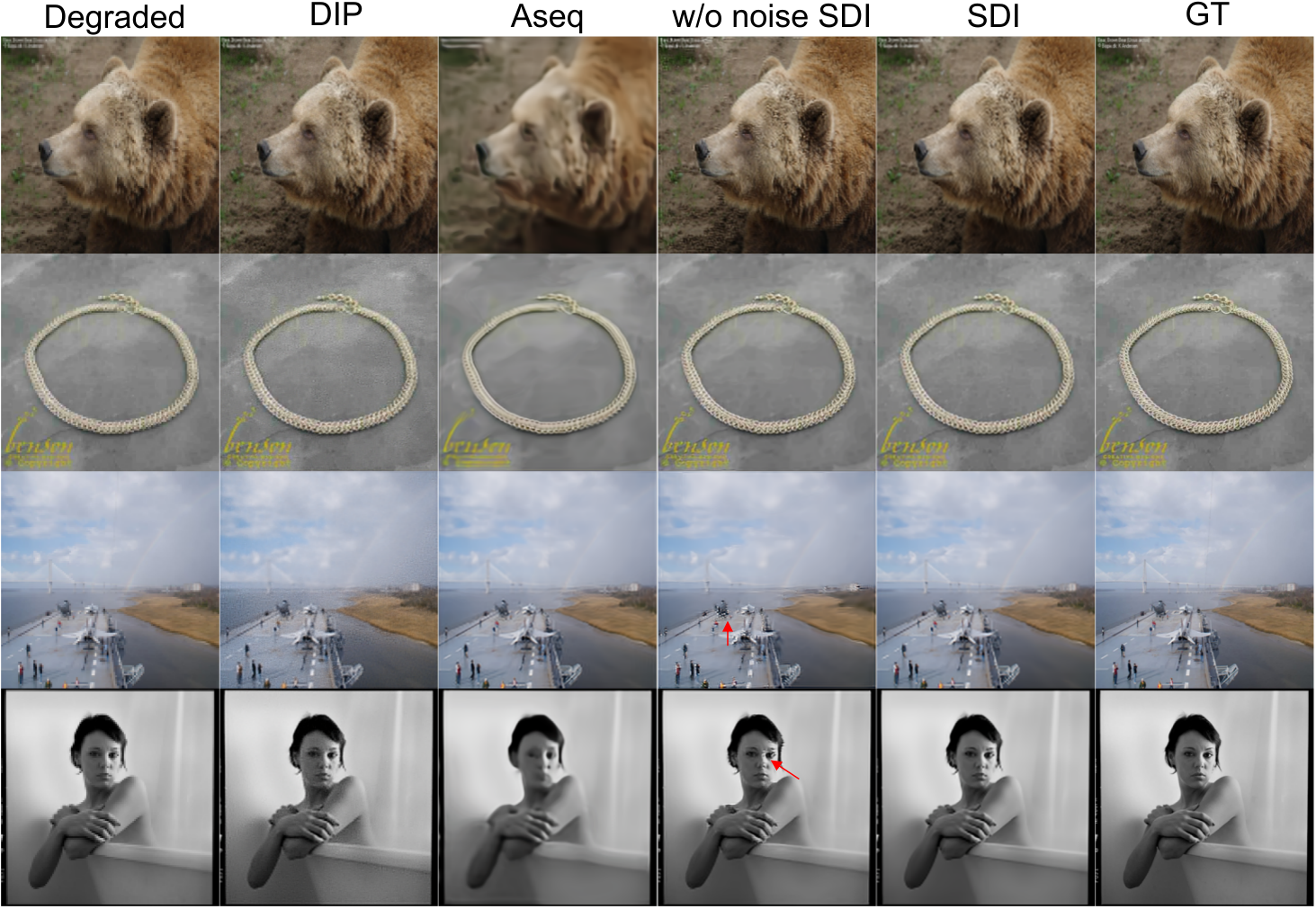}
    \vspace{-1em}
    \caption{Comparison of 2$\times$ super-resolution results across different methods (DIP, Aseq, w/o noise SDI, SDI) against the ground truth (GT), illustrating the enhancement of image details in various scenes including a bear, a necklace, an aircraft carrier, and a person in a bathtub. \label{fig:sr2}}
  \end{figure}
\begin{figure}[H]
  \centering
  \includegraphics[width=\textwidth]{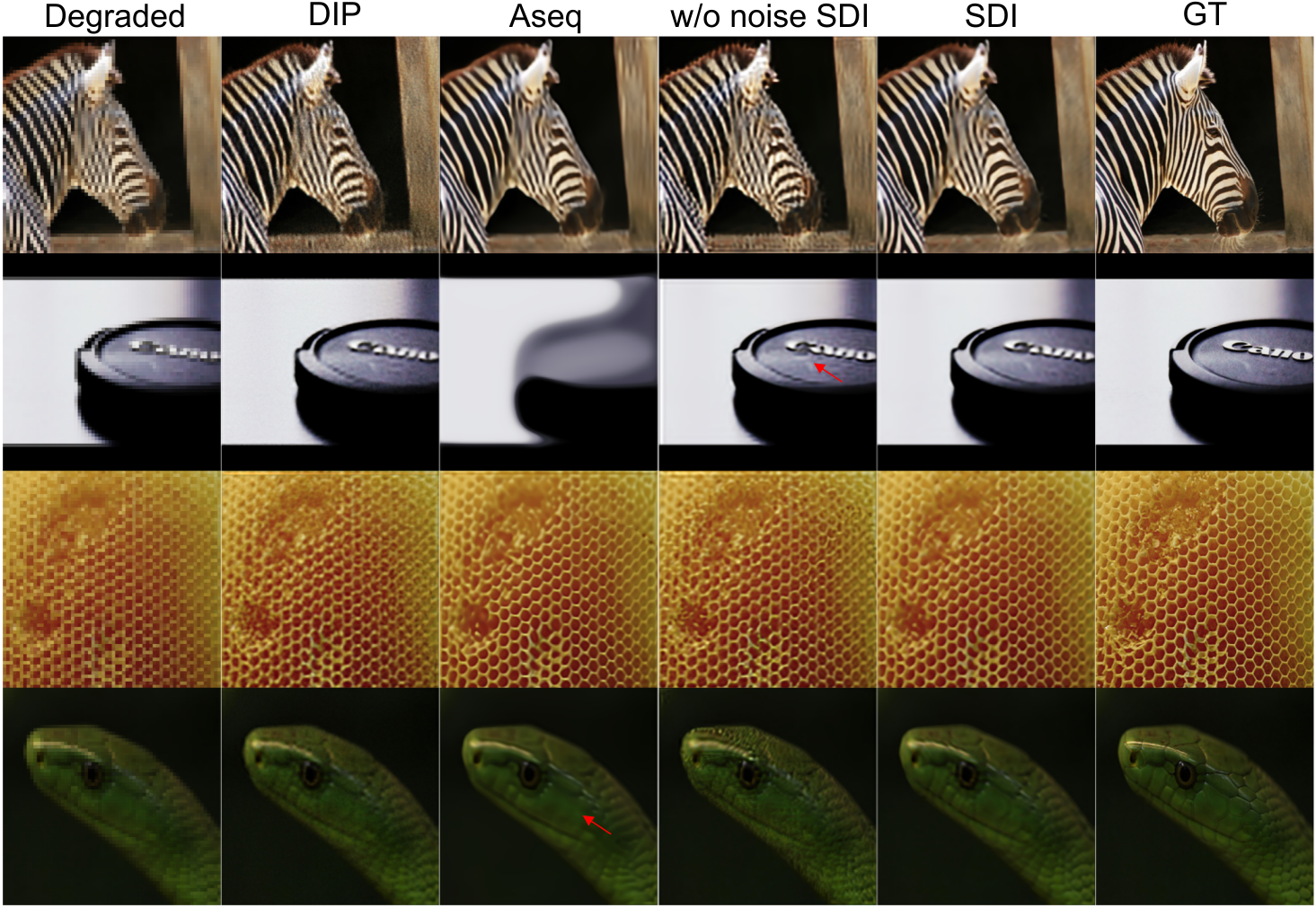}
  \caption{Comparison of 4$\times$ super-resolution results across different methods (DIP, Aseq, w/o noise SDI, SDI) against the ground truth (GT), illustrating the enhancement of fine details in images, including a zebra, a lens lid, a honeycomb, and a snake.\label{fig:sr4}}
\end{figure}

\iftoggle{arxiv}{}{
\newpage
\section*{NeurIPS Paper Checklist}

\begin{enumerate}

\item {\bf Claims}
    \item[] Question: Do the main claims made in the abstract and introduction accurately reflect the paper's contributions and scope?
    \item[] Answer: \answerYes{} 
    \item[] Justification: See abstract and introduction in \Cref{sec:contributions}

\item {\bf Limitations}
    \item[] Question: Does the paper discuss the limitations of the work performed by the authors?
    \item[] Answer: \answerYes{} 
    \item[] Justification: See the discussion in \Cref{sec:limitations}

\item {\bf Theory assumptions and proofs}
    \item[] Question: For each theoretical result, does the paper provide the full set of assumptions and a complete (and correct) proof?
    \item[] Answer: \answerYes{} 
    \item[] Justification: See \Cref{appendix:proof1,appendix:proof2,appendix:proof3}.

    \item {\bf Experimental result reproducibility}
    \item[] Question: Does the paper fully disclose all the information needed to reproduce the main experimental results of the paper to the extent that it affects the main claims and/or conclusions of the paper (regardless of whether the code and data are provided or not)?
    \item[] Answer: \answerYes{} 
    \item[] Justification: See \Cref{sec:simulation,sec:application}

\item {\bf Open access to data and code}
    \item[] Question: Does the paper provide open access to the data and code, with sufficient instructions to faithfully reproduce the main experimental results, as described in supplemental material?
    \item[] Answer: \answerYes{} 
    \item[] Justification: We will make sure that the data and code are open access after review.

\item {\bf Experimental setting/details}
    \item[] Question: Does the paper specify all the training and test details (e.g., data splits, hyperparameters, how they were chosen, type of optimizer, etc.) necessary to understand the results?
    \item[] Answer: \answerYes{} 
    \item[] Justification: See \Cref{sec:simulation,sec:application}

\item {\bf Experiment statistical significance}
    \item[] Question: Does the paper report error bars suitably and correctly defined or other appropriate information about the statistical significance of the experiments?
    \item[] Answer: \answerNA{} 
    \item[] Justification: N/A

\item {\bf Experiments compute resources}
    \item[] Question: For each experiment, does the paper provide sufficient information on the computer resources (type of compute workers, memory, time of execution) needed to reproduce the experiments?
    \item[] Answer: \answerNA{} 
    \item[] Justification: This work is not computation intensive.
    
\item {\bf Code of ethics}
    \item[] Question: Does the research conducted in the paper conform, in every respect, with the NeurIPS Code of Ethics \url{https://neurips.cc/public/EthicsGuidelines}?
    \item[] Answer: \answerYes{} 
    \item[] Justification: We follow the NeurIPS Code of Ethics.

\item {\bf Broader impacts}
    \item[] Question: Does the paper discuss both potential positive societal impacts and negative societal impacts of the work performed?
    \item[] Answer: \answerYes{} 
    \item[] Justification: This work has neutral societal impact.
    
\item {\bf Safeguards}
    \item[] Question: Does the paper describe safeguards that have been put in place for responsible release of data or models that have a high risk for misuse (e.g., pretrained language models, image generators, or scraped datasets)?
    \item[] Answer: \answerNA{} 
    \item[] Justification: This work does not pose a high risk for misuse.

\item {\bf Licenses for existing assets}
    \item[] Question: Are the creators or original owners of assets (e.g., code, data, models), used in the paper, properly credited and are the license and terms of use explicitly mentioned and properly respected?
    \item[] Answer: \answerYes{} 
    \item[] Justification: We will license our code and data under CC-BY 4.0.

\item {\bf New assets}
    \item[] Question: Are new assets introduced in the paper well documented and is the documentation provided alongside the assets?
    \item[] Answer: \answerYes{} 
    \item[] Justification: We will prepare documentation for our code and data.

\item {\bf Crowdsourcing and research with human subjects}
    \item[] Question: For crowdsourcing experiments and research with human subjects, does the paper include the full text of instructions given to participants and screenshots, if applicable, as well as details about compensation (if any)? 
    \item[] Answer: \answerNo{} 
    \item[] Justification: We did not involve crowdsourcing.

\item {\bf Institutional review board (IRB) approvals or equivalent for research with human subjects}
    \item[] Question: Does the paper describe potential risks incurred by study participants, whether such risks were disclosed to the subjects, and whether Institutional Review Board (IRB) approvals (or an equivalent approval/review based on the requirements of your country or institution) were obtained?
    \item[] Answer: \answerNA{} 
    \item[] Justification: Not applicable.

\item {\bf Declaration of LLM usage}
    \item[] Question: Does the paper describe the usage of LLMs if it is an important, original, or non-standard component of the core methods in this research? Note that if the LLM is used only for writing, editing, or formatting purposes and does not impact the core methodology, scientific rigorousness, or originality of the research, declaration is not required.
    \item[] Answer: \answerYes{} 
    \item[] Justification: We use chatgpt for writing, editing, and formatting purposes.

\end{enumerate}
}
\end{document}